 \documentclass[final,5p,times,twocolumn]{elsarticle}

\usepackage{mwe}
\usepackage[english]{babel}
\usepackage{lineno}
\usepackage{amsmath,amsfonts,amssymb,amsthm,epsfig,url,epstopdf,array}
\usepackage{graphicx}
\usepackage[justification=centering]{caption}
\usepackage{multirow}
\usepackage{hyperref}
\usepackage{enumerate}
\usepackage{color}
\usepackage[pdftex,dvipsnames]{xcolor}
\usepackage{float}
\usepackage{subcaption}
\usepackage[flushleft]{threeparttable}
\usepackage{booktabs,caption}
\usepackage{xargs}
\usepackage{pgf}
\usepackage{tikz}
\usepackage{relsize}
\usetikzlibrary{arrows,automata,positioning}
\usepackage[flushleft]{threeparttable} 
\usepackage{algorithm}
\usepackage{algpseudocode}
\usepackage{subcaption}
\hypersetup{pdfauthor={Name}}

\journal{Expert Systems with Applications}

\biboptions{authoryear}
\begin{document}
\begin{frontmatter}

\title{Online learning of windmill time series using Long Short-term Cognitive Networks}


\author[HASSELT]{Alejandro Morales-Hern\'andez}
\address[HASSELT]{Business Informatics Research Group, Hasselt University, Belgium.}

\author[TILBURG]{Gonzalo N\'apoles\corref{mycorrespondingauthor}}
\address[TILBURG]{Department of Cognitive Science \& Artificial Intelligence, Tilburg University, The Netherlands.}

\cortext[mycorrespondingauthor]{Corresponding author}
\ead{g.r.napoles@uvt.nl}

\author[WARSAW]{Agnieszka Jastrzebska}
\address[WARSAW]{Faculty of Mathematics and Information Science, Warsaw University of Technology, Poland.}

\author[TALCA]{Yamisleydi Salgueiro}
\address[TALCA]{Department of Computer Science, Faculty of Engineering, Universidad de Talca, Campus Curic\'o, Chile.}

\author[HASSELT]{Koen Vanhoof}

\begin{abstract}
Forecasting windmill time series is often the basis of other processes such as anomaly detection, health monitoring, or maintenance scheduling. The amount of data generated by windmill farms makes online learning the most viable strategy to follow. Such settings require retraining the model each time a new batch of data is available. However, updating the model with new information is often very expensive when using traditional Recurrent Neural Networks (RNNs). In this paper, we use Long Short-term Cognitive Networks (LSTCNs) to forecast windmill time series in online settings. These recently introduced neural systems consist of chained Short-term Cognitive Network blocks, each processing a temporal data chunk. The learning algorithm of these blocks is based on a very fast, deterministic learning rule that makes LSTCNs suitable for online learning tasks. The numerical simulations using a case study involving four windmills showed that our approach reported the lowest forecasting errors with respect to a simple RNN, a Long Short-term Memory, a Gated Recurrent Unit, and a Hidden Markov Model. What is perhaps more important is that the LSTCN approach is significantly faster than these state-of-the-art models.
\end{abstract}


\end{frontmatter}


\section{Introduction}
\label{sec:introduction}

Humanity's sustainable development requires the adoption of less environmentally aggressive energy sources. Over the last years, renewable energy sources (RES) have increased their presence in the energy matrix of several countries. These clean energies are less polluting, renewable, and abundant in nature. However, limitations such as volatility and intermittency reduce their reliability and stability for power systems. This hinders the integration of renewable sources into the main grid and increases their generation costs \citep{Sinsel2020}.

Power generation forecasting \citep{foley2012current} is one of the approaches adopted to facilitate the optimal integration of RES in power systems. Overall, the goal of power generation forecasting is to know in advance the possible disparity between generation and demand due to fluctuations in energy sources \citep{Ahmed2019}. Forecasting methods used for renewable energies are based on physical, statistical, or machine learning (ML) models. Although ML models often achieve the highest performance compared to other models, their deployment in real applications is limited \citep{Jorgensen2020}. On the one hand, most ML models require feature engineering before building the model and lack interpretability. On the other hand, these methods usually assume that the training data is completely available in advance. Hence, most ML methods are unable to incorporate new information into the previously constructed models \citep{Wang2019a}. 

Within clean energy approaches, wind energy has shown sustained growth in installed capacity and exploitation in recent years \citep{Ahmed2020}. However, wind energy involves some peculiarities to be considered when designing new forecasting solutions. Firstly, wind-based power generation can heavily be affected by weather variability, which means that the power generation fluctuates with extreme weather phenomena (i.e., frontal systems or rapidly evolving low-pressure systems). Weather events are unavoidable, but their impact can be minimized if anticipated in advance. Secondly, wind generators are dynamic systems that behave differently over time (i.e., due to wear of turbine components, maintenance, etc). 
Finally, the data generated by windmills is not static since they will continue to operate, thus producing new pieces of data. These characteristics make traditional ML methods inadequate to model the dynamics of these systems properly. This means that new approaches are needed to improve the prediction of wind generation. The development of algorithms capable of learning beyond the production phase will also allow them to be kept up-to-date at all times \citep{Losing2018}.

\vspace{-0.5mm}

Recently, \cite{Napoles2021lstcn} introduced a recurrent neural system termed \textit{Long Short-term Cognitive Network} (LSTCN) that seems suitable for online learning setting where data might be volatile. Moreover, the cognitive component of such a recurrent neural network allows for interpretability and it is given by two facts. Firstly, neural concepts and weights have a well-defined meaning for the problem domain being modeled. This means that the resulting model can easily be interpreted with little effort. For example, in \citep{Napoles2021lstcn} the authors discussed a measure to compute the relevance of each variable in multivariate time series without the need for any post-hoc method. Secondly, the domain expert can insert knowledge into the network by modifying the prior knowledge matrix, which is not altered during the learning process. For example, we can modify some connections in the weight matrix to manually encode patterns that have not yet been observed in the data or that were observed under exceptional circumstances.

Despite the advantages of the LSTCN model when it comes to its forecasting capabilities, intrinsic interpretability and short training time, it has not yet been applied to a real-world problem, as far as we know. In addition, we have little knowledge of the performance of this brand new model on online learning settings operating with volatile data that might be shortly available. Such a lack of knowledge and the challenges related to the wind prediction described above have motivated us to study the LSTCNs' performance on a real-world problem concerning the power forecasting of four windmills. 

More explicitly, this paper elaborates on the task of forecasting power generation in windmills using the LSTCN model. By doing that, we propose an LSTCN-based pipeline to tackle the related online learning problem where each data chuck is processed only once. In this pipeline, each iteration processes a data chunk using a Short-term Cognitive Network (STCN) block \citep{Napoles-stcn} that operates with the knowledge transferred from the previous block. This means that the model can be retrained without compromising what the network has learned from previous data chunks. The numerical simulations show that our solution (i) outperforms state-of-the-art recurrent neural networks when it comes to the forecasting error and (ii) reports significantly shorter training and test times.

The remainder of the paper is organized as follows. Section \ref{sec:literature} revises the literature about recurrent neural networks used to forecast windmill time series. Section \ref{sec:lstcn} presents the proposed LSTCN-based power forecasting model for an online learning setting. Section \ref{sec:simulations} describes the case study, the state-of-the-art recurrent models used for comparison purposes and the simulation results. Finally, Section \ref{sec:remarks} concludes the paper and suggests further research directions to be explored.

\section{Forecasting models with recurrent neural networks}
\label{sec:literature}

Neural networks are a family of biology-inspired computational models that have found applications in many fields. An example of engineering applications of neural models is the support of wind turbine operation and maintenance. In this area, neural models dedicated to the analysis of temporal data have proven to be quite useful. This is motivated by the fact that typical data describing the operation of a wind turbine are collected by sensors forming a supervisory control and data acquisition (SCADA) system~\citep{Du2017,Weerakody2021}. Such data come in the form of long sequences of numerical values, thus making Recurrent Neural Networks (RNNs) the right choice for processing such data. This section briefly revises the literature on the applications of RNNs on data analysis in the area of wind turbine operation and maintenance support. 

RNNs differ from other neural networks in the way the input data is propagated. In standard neural networks, the input data is processed in a feed-forward manner, meaning the signal is transmitted unidirectionally. In RNN models, the signal goes through neurons that can have backward connections from further layers to earlier layers~\citep{Che2018}. Depending on a particular neural model architecture, we can restrict the layers with feedback connections to only selected ones. The overall idea is to allow the network to ``revisit'' nodes, which mimics the natural phenomenon of memory~\citep{Kong2019}. RNNs turned out to be useful for accurate time series prediction tasks~\citep{Strobelt2018}, including wind turbine time series prediction~\citep{Cui2021}.

As reported by \cite{Zhang2020}, the task of analyzing wind turbine data often involves building a regression model operating on multi-attribute data from SCADA sensors. Such models can help us understand the data \citep{Delgado2021,Janssens2016}.

Currently, the most popular variant of RNN in the field of wind turbine data processing is the Long Short-Term Memory (LSTM) model~\citep{Hochreiter1997,Mishra2020}. In this model, the inner operations are defined by neural gates called \textit{cell}, \textit{input gate}, \textit{output gate}, and \textit{forget gate}. The cell acts as the memory, while the other components determine the way the signal propagates through the neural architecture \citep{Zhang2018}. The introduction of these specialized units helped prevent (to some extent) the gradient problems associated with training RNN models~\citep{Sherstinsky2020}.

Existing neural network approaches to wind turbine data forecasting do not pay enough attention to the issue of model complexity and efficiency. In most studies, authors reduce the available set of input variables rather than optimizing the neural architecture used. For example, \cite{Feng2019} used the LSTM model with hand-picked three SCADA input variables, while \cite{Riganti2018} used eleven SCADA variables. \cite{Qian2019} also used LSTM to predict wind turbine data. In their study, the initial set of input variables consisted of 121 series, but this was later reduced to only three variables and then to two variables using the Mahalanobis distance method. The issue of preprocessing and feature selection was also raised by \cite{Wang2018}, suggesting Principal Component Analysis to reduce the dimensionality of the data. 

LSTM has been found to perform well even when the time series variables are of incompatible types. It is worth citing the study of \cite{Lei2019}, who used LSTM to predict two qualitatively different types of time series simultaneously: (i) vibration measurements that have a high sampling rate and (ii) slow varying measurements (e.g., bearing temperature). It should be noted that existing studies bring additional techniques that enhance the capabilities of the standard LSTM model. For example, \cite{Cao2019} propose segmenting the data and using segment-related features instead of raw signals. \cite{Xiang2021} also do not use raw signals. Instead, they use Convolutional Neural Networks (CNNs) to extract the dynamic features of the data, which is then fed to LSTM. A similar approach, combining CNN with LSTM, was presented by \cite{Xue2021}. Another interesting technique was introduced by \cite{Chen2021}, who combined LSTM with an auto-encoder (AE) neural network so that their model can detect and reject anomalies while achieving better results for non-anomalous data. \cite{Liu2020} used wavelet decomposition together with LSTM and found that it achieves better results than standard LSTM, but this comes at the cost of increased time complexity (training time increases by about 30\%). Other studies on LSTM and wind power prediction have focused on tuning the LSTM architecture, for example, by testing different transformation functions \citep{Yin2020} or by adding a specialized imputation module for missing data \citep{Li2019}.

In addition, the bidirectional LSTM model \citep{gers2002learning} has also been applied to forecast wind turbine data. The application of this model was found in the study of \cite{Zhen2020} and, in a deeper architecture, in the study of \cite{Cao2019b}.

While most of the recently published studies using neural models to predict multivariate wind turbine time series employ LSTM, there are also several alternative approaches focusing on other RNN variants. For example, there are several papers on the use of Elman neural networks in forecasting multivariate wind turbine data \citep{Lin2013,Lin2016}. \cite{Kramti2018} also applied Elman neural networks, but using a slightly modified architecture. Likewise, we should mention the work of \cite{Lopez2020}, which involved Echo State Network and LSTM. Finally, it is worth mentioning the work of \cite{Kong2020}, in which the task of processing data from wind turbines is implemented using CNNs and Gated Recurrent Unit (GRU) \citep{cho-etal-2014-learning}. The latter neural architecture is a variant of RNN, which can be seen as a simplification of the LSTM architecture. GRU was also used in the study of \cite{Niu2020}, which employs attention mechanisms to reduce the forecasting error. 

There are other models equipped with reasoning mechanisms similar to the one used by neural networks. In particular, the concept of ``neuron" can also be found in Hidden Markov Models (HMMs) \citep{rabiner1989tutorial}. Such neurons are implemented as \textit{states}, and the set of states essentially plays a role analogous to that of hidden neurons in a standard neural network. HMMs have also found applications in wind power forecasting. The studies of \cite{Bhaumik2019} and \cite{Qu2021} should be mentioned in this context. Both research teams highlight decent predictions and robustness to noise in the data. 

As pointed out by \cite{Manero2018}, the task of comparing wind energy forecasting approaches described in the literature is challenging due to several factors such as the differences in time series datasets, the alternative forecast horizons, etc. In this paper, we will conduct experiments for key state-of-the-art models for our data alongside the LSTCN formalism. The methodology adopted allows us to draw conclusions about the forecasting accuracy of different models and compare their empirical computational complexity.

\section{Long Short-term Cognitive Network}
\label{sec:lstcn}

This section elaborates on the LSTCN model used for online learning of multivariate time series. The first sub-section will explain how to prepare the data to simulate an online learning problem, while the remaining ones will introduce the network architecture and the learning algorithm.

\subsection{Data preparation for online learning simulations}
\label{sec:lstcn:notation_data}

Let $x \in \mathbb{R}$ be a variable observed over a discrete time scale within a period $t \in \{1, 2, \ldots, T\}$ where $T \in \mathbb{N}$ is the number of observations. Hence, a univariate time series can be defined as a sequence of observations $\{x^{(t)}\}_{t=1}^{T} = \{x^{(1)}, x^{(2)}, \ldots, x^{(T)}\}$. Similarly, we can define a multivariate time series as a sequence $\{X^{(t)}\}_{t=1}^{T} =$ $\{ X^{(1)}, X^{(2)}, \ldots, X^{(T)} \}$ of vectors of $M$ variables, such that $X^{(t)} = [x_{1}^{(t)}, x_{2}^{(t)}, \ldots, x_{M}^{(t)} ]$. A model \textit{F} is used to forecast the next $L<T$ steps ahead. In this paper, we assume that the model \textit{F} is built as a sequence of neural blocks with local learning capabilities, each able to capture the trends in the current time patch (i.e., a chunk of the time series) being processed. Both the network architecture and the parameter learning algorithm will be detailed in the following sub-sections.

\begin{figure}[!htbp]
\centering
\begin{subfigure}[b]{0.48\textwidth}
\centering
\includegraphics[width=8cm]{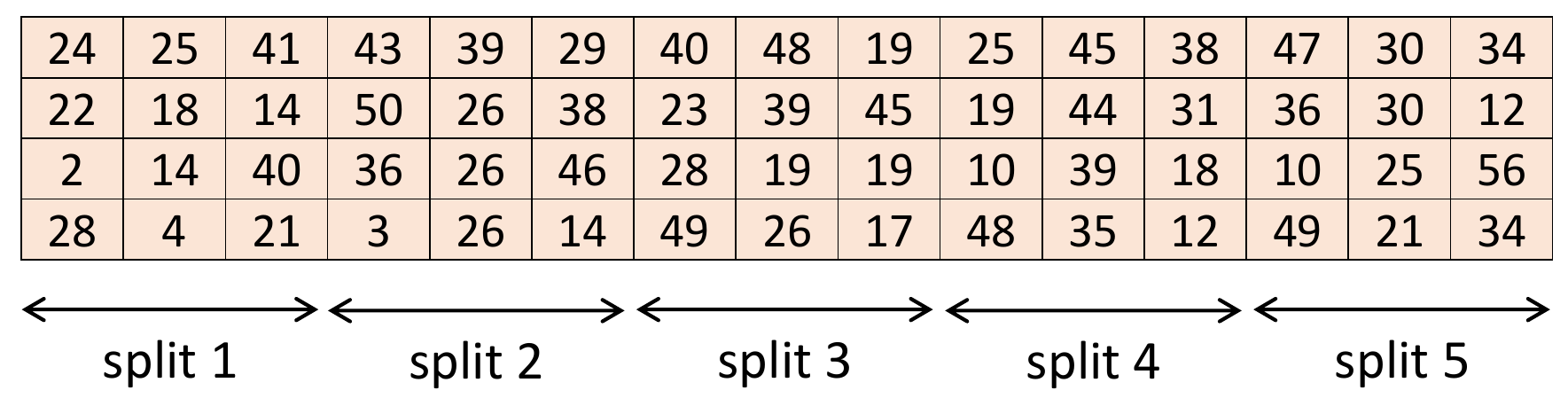}
\caption{Original multivariate time series\vspace{1mm}}
\label{fig:preprocessingA}
\end{subfigure}
\vspace{1mm}
\begin{subfigure}[b]{0.48\textwidth}
\centering
\includegraphics[width=9cm]{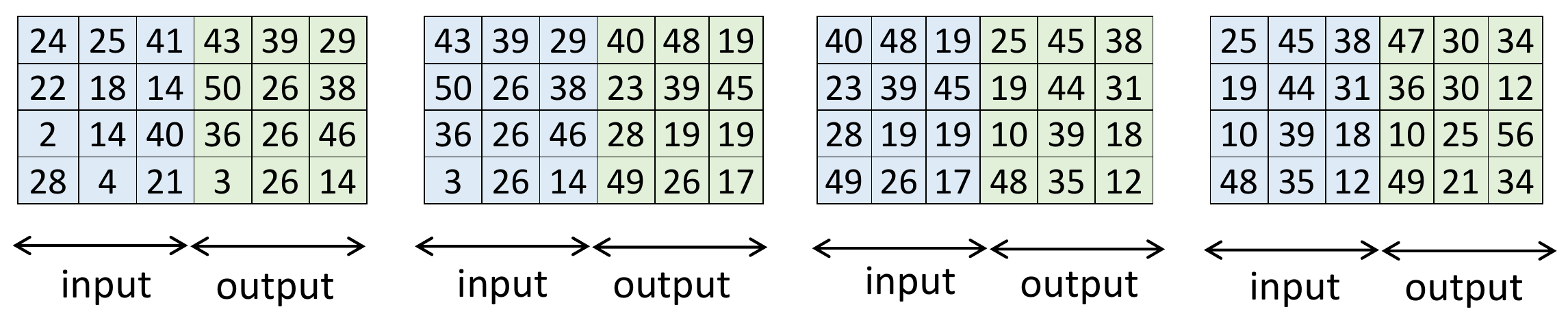}
\caption{Rolling mean}
\label{fig:preprocessingB}
\end{subfigure}
\begin{subfigure}[b]{0.48\textwidth}
\centering
\includegraphics[width=9cm]{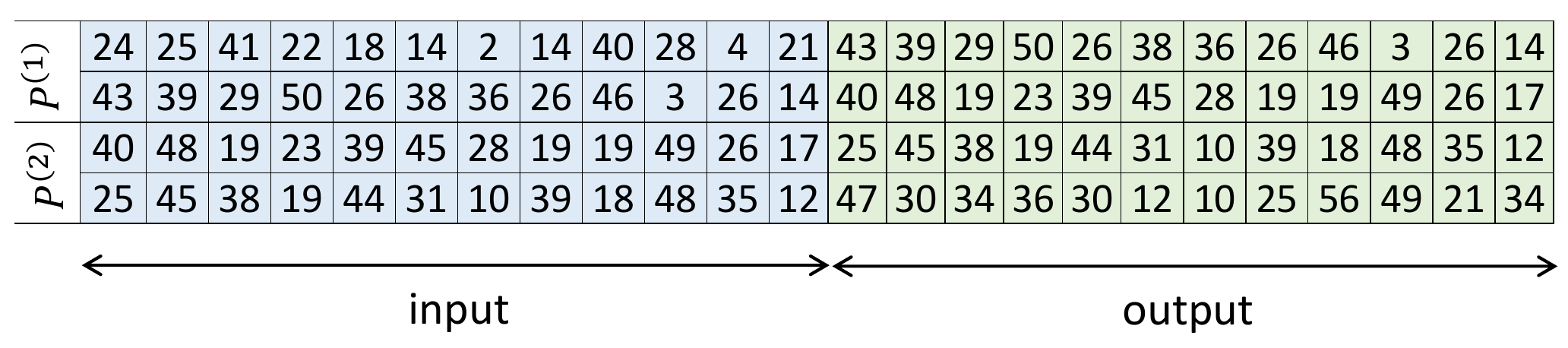}
\caption{Flattening}
\label{fig:preprocessingC}
\end{subfigure}

\caption{Data pre-processing using $R=L=3$. (a) The original multivariate time series $X\in \mathbb{R}^{M \times T}$, with rows as variables and columns as timestamps. (b) Selection of sub-sequences of the time series according to parameters $R$ and $L$. (c) Each sub-sequence is flattened to obtain the temporal instances. In this example, the flattened dataset is divided into two time parches.}
\label{fig:preprocessing}
\end{figure}

Let us assume that $X \in \mathbb{R}^{M \times T}$ is a dataset comprising a multivariate time series (Figure \ref{fig:preprocessingA}). Firstly, we need to transform $X$ into a set of $Q$ tuples with the form $(X^{(t-R)}, X^{(t+L)}), t-R>0, t+L \leq T$ where $R$ represents how many past steps we will use to forecast the following $L$ steps ahead (see Figure \ref{fig:preprocessingB}). In this paper, we assume that $R=L$ for the sake of simplicity. Secondly, each component in the tuple is flattened such that we obtain a $Q \times (M(R+L))$ matrix. Finally, we create a partition $P = \{P^{(1)},\ldots,P^{(k)},\ldots,P^{(K)} \}$ from the set of flattened tuples such that $P^{(k)} = (P_1^{(k)}, P_2^{(k)})$ is the $k$-th time patch involving two data pieces $P_1^{(k)}, P_2^{(k)} \in \mathbb{R}^{C \times N}$, where $N=M R$ and $C$ denotes the number of instances in that time patch.

Figure \ref{fig:preprocessing} shows an example of such a pre-processing method. First, the times series is split into chunks of equal length as defined by the $L$ and $R$ parameters. Second, we use the resulting chunks to create a set of input-output pairs. Finally, we flatten these pairs to obtain the tuples with the inputs to the network and the corresponding expected outputs.

It should be highlighted that the forecasting model will have access to a time patch in each iteration, as it usually happens in an online scenario. If the neural model is fed with several time steps, then it will be able to forecast multiple-step ahead of all variables describing the time series.

\subsection{Network architecture and neural reasoning}
\label{sec:lstcn:model}

In the online learning setting, we consider a time series (regardless of the number of observed variables) as a sequence of time patches of a certain length. Such a sequence refers to the partition $P = \{P^{(1)},\ldots,P^{(k)},\ldots,P^{(K)} \}$ obtained with the data preparation steps discussed in the previous subsection. Hence, the proposed network architecture consists of an LSTCN model able to process the sequence of time patches.

\begin{figure}[!htbp]
\center
\includegraphics[width=9cm]{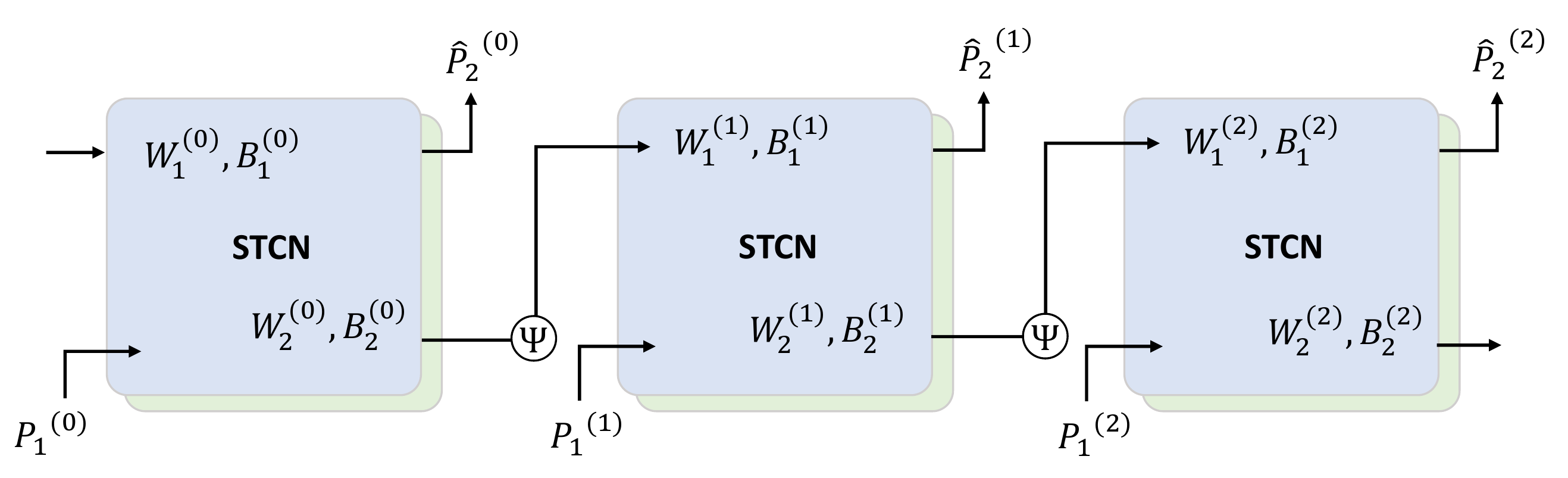}
\caption{LSTCN architecture of three STCN blocks. The weights learned in the current block are transferred to the following STCN block as a prior knowledge matrix.}
\label{fig:lstcn}
\end{figure}

\begin{figure}[!htbp]
\center
\includegraphics[width=8cm]{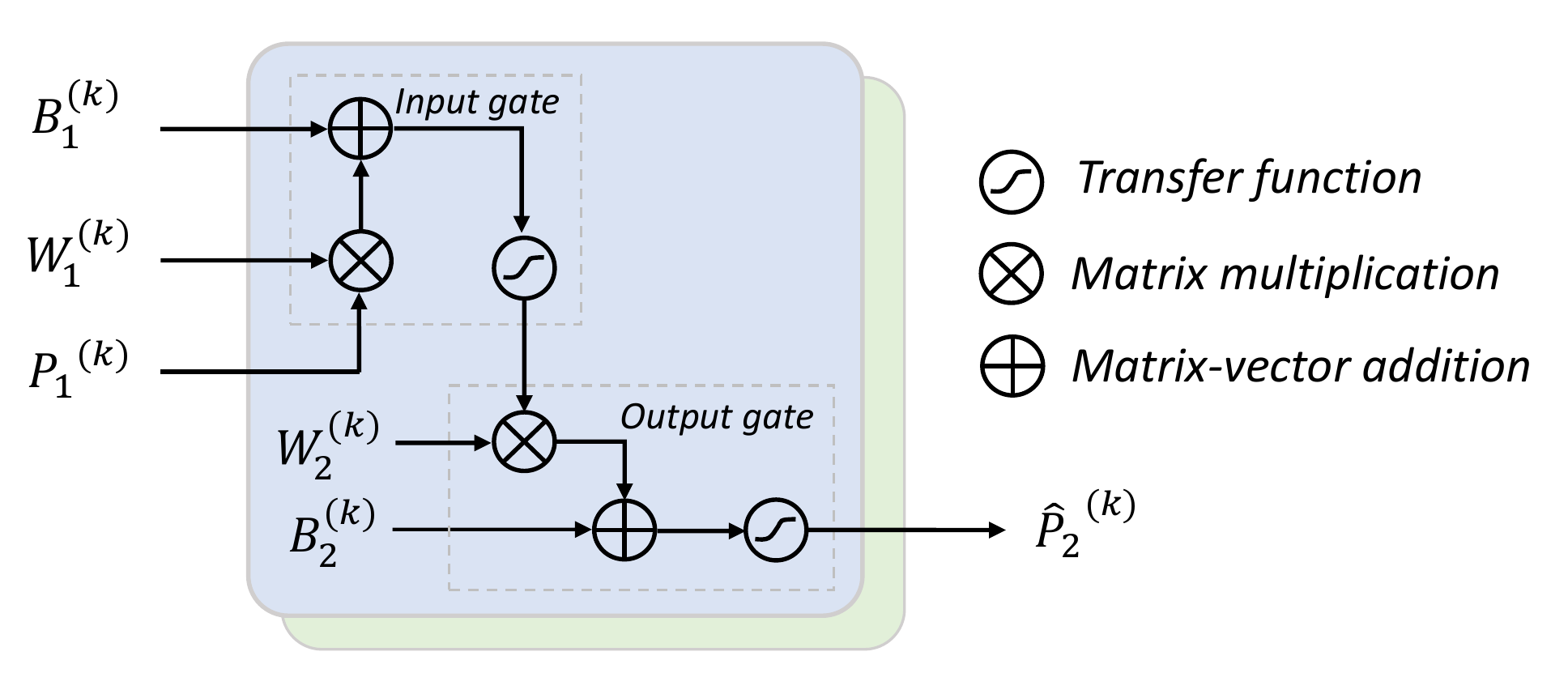}
\caption{Reasoning within an STCN block. Firstly, the current time patch is mixed with the prior knowledge matrices $W_{1}^{(k)}$ and $B_{1}^{(k)}$. This operation produces a temporal state matrix $H^{(k)}$. Secondly, we operate the $H^{(k)}$ matrix with the matrices $W_{2}^{(k)}$ and $B_{2}^{(k)}$. The result of such an operation will be an approximation of the expected output $P_2^{(k)}$.}
\label{fig:stcn2}
\end{figure}

An LSTCN model can be defined as a collection of STCN blocks, each processing a specific time patch and transferring knowledge to the following STCN block in the form of weight matrices. Figure \ref{fig:lstcn} shows the recurrent pipeline of an LSTCN involving three STCN blocks to model a multivariate time series decomposed into three time patches. It should be highlighted that learning happens inside each STCN block to prevent the information flow from vanishing as the network processes more time patches. Moreover, weights estimated in the current STCN block are transferred to the following STCN block to perform the next reasoning process (see Figure \ref{fig:stcn2}). These weights will no longer be modified in subsequent learning processes, which allow preserving the knowledge we have learned up to the current time patch. That makes our approach suitable for the online learning setting.

The reasoning within an STCN block involves two gates: the \textit{input gate} and the \textit{output gate}. The input gate operates the prior knowledge matrix $W_1^{(k)} \in \mathbb{R}^{N \times N}$ with the input data $P_1^{(k)} \in \mathbb{R}^{C \times N}$ and the prior bias matrix $B_1^{(k)} \in \mathbb{R}^{1 \times N}$ denoting the bias weights. Both matrices $W_1^{(k)}$ and $B_1^{(k)}$ are transferred from the previous block and remain locked during the learning phase to be performed in that STCN block. The result of the input gate is a temporal state $H^{(k)} \in \mathbb{R}^{C \times N}$ that represents the outcome that the block would have produced given $P_1^{(k)}$ if the block would not have been adjusted to the block's expected output $P_2^{(k)}$. Such an adaptation is done in the output gate where the temporal state is operated with the matrices $W_2^{(k)} \in \mathbb{R}^{N \times N}$ and $B_2^{(k)} \in \mathbb{R}^{1 \times N}$, which contain learnable weights. Figure \ref{fig:stcn2} depicts the reasoning process within the $k$-th block.

Equations \eqref{eq:stcn1} and \eqref{eq:stcn2} show the short-term reasoning process of this model in the $k$-th iteration,

\begin{equation}
\label{eq:stcn1}
\hat{P_2}^{(k)}=f\left(H^{(k)} W_2^{(k)} \oplus B_2^{(k)}\right)
\end{equation}
\noindent and
\begin{equation}
\label{eq:stcn2}
H^{(k)}=f\left(P_1^{(k)} W_1^{(k)} \oplus B_1^{(k)} \right)
\end{equation}

\noindent where $f(x) = \frac{1}{1+e^{-x}}$, whereas $\hat{P_2}^{(k)}$ is an approximation of the expected block's output. In these equations, the $\oplus$ operator performs a matrix-vector addition by operating each row of a given matrix with a vector, provided that both the matrix and the vector have the same number of columns. Notice that we assumed that values to be forecast are in the $[0,1]$ interval.

As mentioned, the LSTCN model consists of a sequential collection of STCN blocks. In this neural system, the knowledge from one block is passed to the next one using an aggregation procedure (see Figure \ref{fig:lstcn}). This aggregation operates on the knowledge learned in the previous block (that is to say, the $W_2^{(k-1)}$ matrix). In this paper, we use the following non-linear operator in all our simulations:

\begin{equation} \label{eq:aggregation1}
W_1^{(k)} = \Psi(W_2^{(k-1)}), k-1 \geq 0
\end{equation}
\noindent and
\begin{equation} \label{eq:aggregation2}
B_1^{(k)} = \Psi(B_2^{(k-1)}), k-1 \geq 0
\end{equation}

\noindent such that $\Psi(x) = tanh(x)$. However, we can design operators combining the knowledge in both $W_1^{(k-1)}$ and $W_2^{(k-1)}$.

There is an important detail to be discussed. Once we have processed the available sequence (i.e., performed $K$ short-term reasoning steps with their corresponding learning processes), the whole LSTCN model will narrow down to the last STCN block. Therefore, that network will be used to forecast new data chunks as they arrive and a new learning process will follow, as needed in online learning settings. 

\subsection{Parameter learning}
\label{sec:lstcn:learning}

Training the LSTCN in Figure \ref{fig:lstcn} means training each STCN block with its corresponding time patch. The learning process within a block is partially independent of other blocks as it only uses the prior weights matrices that are transferred from the previous block. As mentioned, these prior knowledge matrices are used to compute the temporal state and are not modified during the block's learning process.

The learning task within an STCN block can be summarized as follows. Given a temporal state $H^{(k)}$ resulting from the input gate and the block's expected output $P_2^{(k)}$, we need to compute the matrices $W_2^{(k)} \in \mathbb{R}^{N \times N}$ and $B_2^{(k)} \in \mathbb{R}^{1 \times N}$. 


Mathematically speaking, the learning is performed by solving a system of linear equations that adapt the temporal state to the expected output. Equation \eqref{eq:ridge} displays the deterministic learning rule solving this regression problem,

\begin{equation}
\label{eq:ridge}
\begin{bmatrix} W_2^{(k)} \\ B_2^{(k)} \end{bmatrix} = \left( \left( \Phi^{(k)} \right)^{\top} \Phi^{(k)} + \lambda \Omega^{(k)} \right)^{-1} \left( \Phi^{(k)} \right)^{\top} f^{-} \left(P_2^{(k)}\right)
\end{equation}

\noindent where $\Phi^{(k)}=(H^{(k)}|A)$ such that $A_{C \times 1}$ is a column vector filled with ones, $\Omega^{(k)}$ denotes the diagonal matrix of $(\Phi^{(k)})^{\top} \Phi^{(k)}$, while $\lambda \geq 0$ denotes the ridge regularization penalty. This learning rule assumes that the neuron's activation values inner layer are standardized. When the final weights are returned, they are adjusted back into their original scale.

It shall be noted that we need to specify $W_1^{(0)}$ and $B_1^{(0)}$ in the first STCN block. We can use a transfer learning approach from a previous learning process or it can be provided by domain experts. Since this information is not available, we fit a single STCN block without an intermediate state (i.e., $H^{(0)}=P_1^{(0)}$) on a smoothed representation of the whole (available) time series. The smoothed time series is obtained using the moving average method for a given window size.

\begin{figure*}[!htbp]
    \centering
    \includegraphics[width=14.5cm]{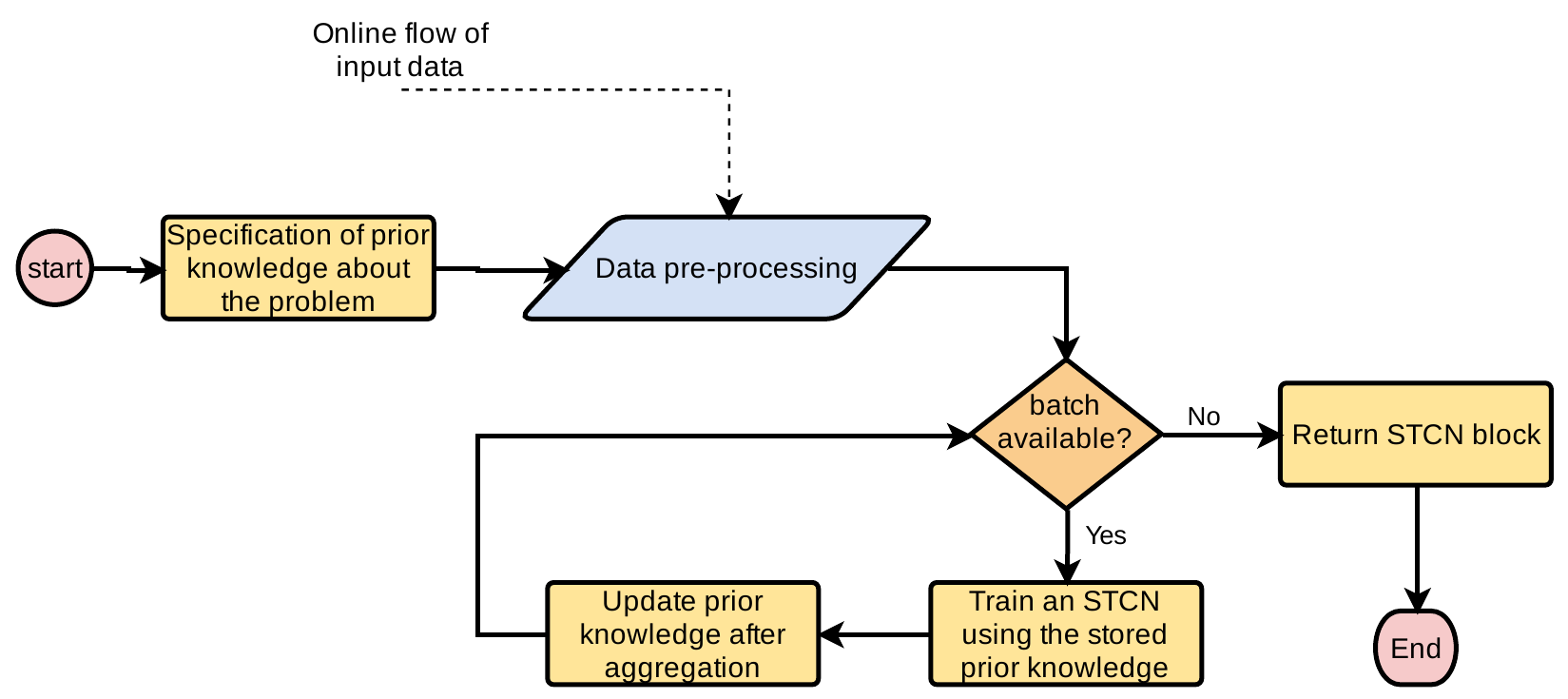}
    \caption{The LSTCN model can be seen as a sequential collection of STCN blocks that perform iterative learning. When a new chunk of data is available, a new STCN block is trained and the prior knowledge is updated using an aggregation procedure.}
    \label{fig:flow_chart}
\end{figure*}

Figure \ref{fig:flow_chart} portrays the workflow of the iterative learning process of an LSTCN model. An incoming chunk of data triggers a new training process on the last STCN block using the stored knowledge that the network has learned in previous iterations. After that, the prior knowledge matrices are recomputed using an aggregation operator and stored to be used as prior knowledge when performing reasoning.

\section{Numerical simulations}
\label{sec:simulations}

In this section, we will explore the performance (forecasting error and training time) of the proposed LSTCN-based online forecasting model for windmill time series.

\subsection{Description of windmill datasets}
\label{sec:simulations:datasets}

To conduct our experiments, we adopted four public datasets from the ENGIE web page\footnote{\href{}{https://opendata-renewables.engie.com/explore/index}}. Each dataset corresponds to a windmill where measurements were recorded every 10 minutes from 2013 to 2017. The time series of each windmill contains 264,671 timestamps. Eight variables concerning the windmill and environmental conditions were selected: \textit{generated power, rotor temperature, rotor bearing temperature, gearbox inlet temperature, generator stator temperature, wind speed, outdoor temperature,} and \textit{nacelle temperature}.

As of the pre-processing steps, we removed duplicated timestamps, imputed missing timestamps and values, and applied a min-max normalization. Moreover, the data preparation procedure described in Figure \ref{fig:preprocessing} was applied to each dataset. Table \ref{tab:datasets} displays a descriptive summary of all datasets after normalization where the minimum, median and maximum of the absolute Pearson's correlation values among the variables are denoted as $min$, $med$, $max$, respectively.

\begin{table}[!htbp]
\center
\small
\caption{Descriptive statistics for the windmill datasets}
\begin{tabular}{c|rrr}
\hline
Dataset & $min$ & $med$ & $max$ \\
\hline
1 & 0.0708 & 0.2799 & 0.9456\\ \hline			
2 & 0.0888 & 0.3032 & 0.8848\\ \hline
3 & 0.0687 & 0.3014 & 0.9497\\ \hline
4 & 0.0835 & 0.3148 & 0.9441\\ \hline
\end{tabular}							
\label{tab:datasets}
\end{table}

We split each dataset using a hold-out approach (80\% for training and 20\% for testing purposes). As for the performance metric, we use the mean absolute error (MAE) in all simulations reported in this section. In addition, we report the training and test times of each forecasting model. The training time (in seconds) of each algorithm was computed by adding the time needed to train the algorithm in each time patch. Finally, we arbitrarily fix the patch size to 1024.

\subsection{Recurrent online learning models}
\label{sec:simulations:algorithms}

We contrast the LSTCNs' performance against four recurrent learning networks used to handle online learning settings. The models adopted for comparison are GRU, LSTM, HMM, and a fully connected Recurrent Neural Network (RNN) where the output is to be fed back to the input.

The RNN, LSTM and GRU networks were implemented using Keras v2.4.3, while HMM was implemented using the \textit{hmmlearn} library\footnote{\url{https://github.com/hmmlearn/hmmlearn}}. The training of these models was adapted to online learning scenarios. In practice, this means that RNN, GRU, and LSTM were retrained on each time patch using the prior knowledge structures learned in previous learning steps. In the HMM-based forecasting model, the transition probability matrix is passed from one patch to another, and it is updated based on the new information.

In the LSTCN model, we used $L=\{6, 48, 72\}$ such that $R=L$ (hereinafter we will only refer to $L$) and $w=10$. Notice that given the sampling interval of the data, six steps represent one hour while 72 steps represent half a day. We did not perform parameter tuning since the online learning setting demands fast re-training of these recurrent models when a new data chunk arrives. It would not be feasible to fine-tune the hyperparameters in each iteration since such a process is computationally demanding. Instead, we retained the default parameters reported on the corresponding Keras layers. In the HMM-based model, we used four hidden states and Gaussian emissions to generate the predictions. These parameter values were arbitrarily selected without further experimentation.

\subsection{Results and discussion}
\label{sec:simulations:results}

\begin{figure*}[!ht]
\centering

\begin{subfigure}[b]{0.49\textwidth}
\centering
\includegraphics[width=\textwidth]{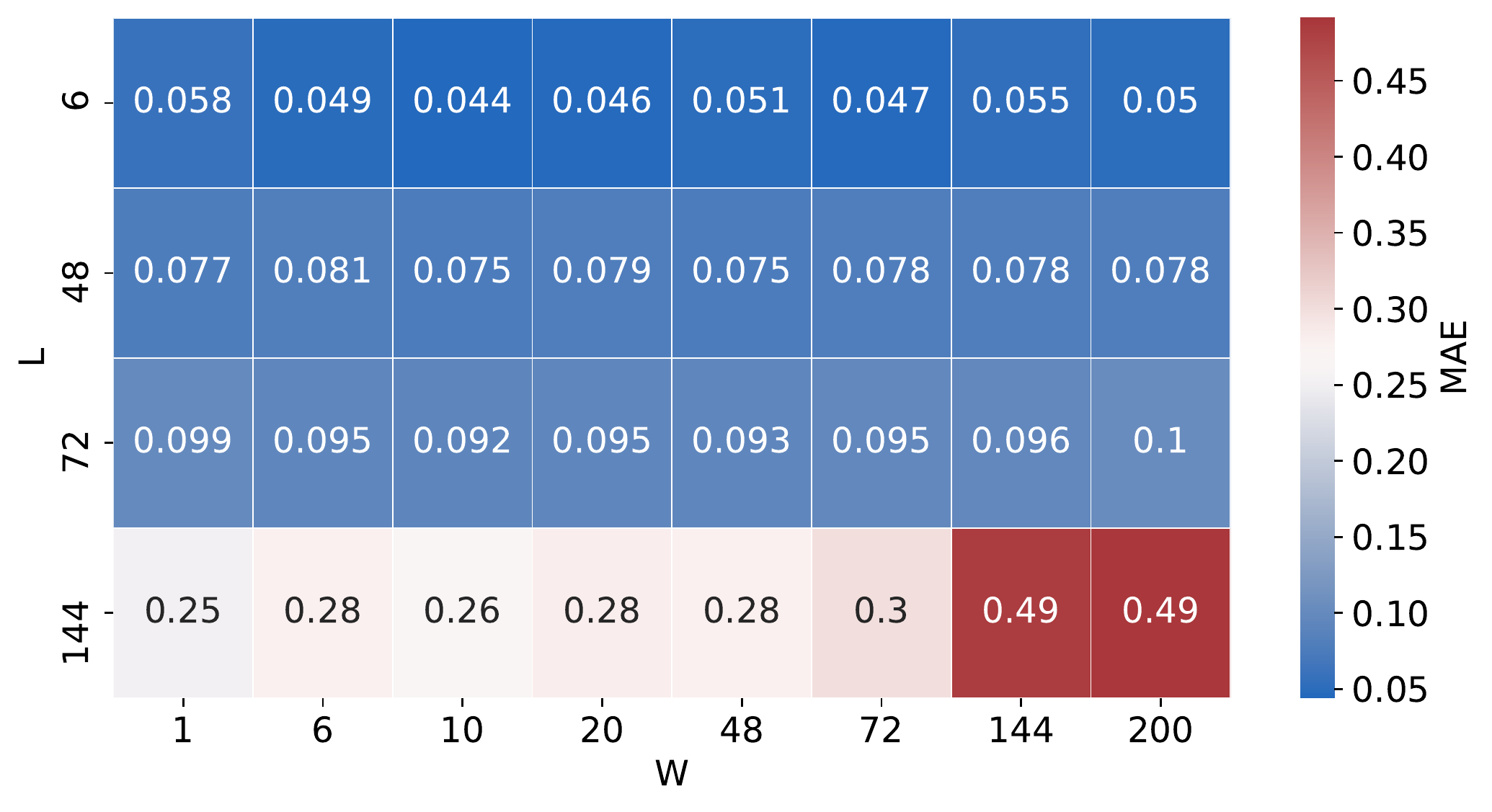}
\caption{WT1}
\label{fig:opt_lw_wt1}
\end{subfigure}
\begin{subfigure}[b]{0.49\textwidth}
\centering
\includegraphics[width=\textwidth]{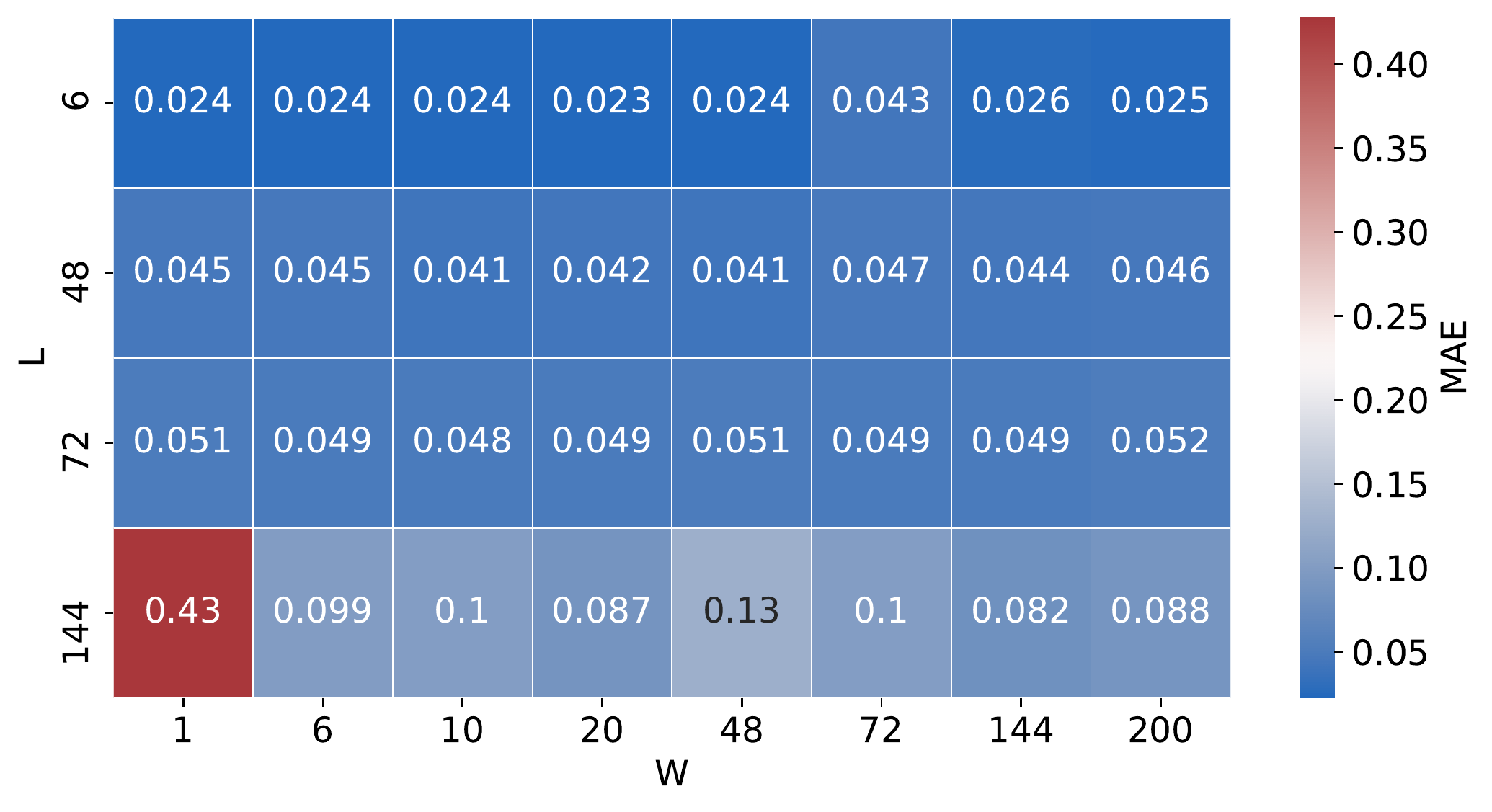}
\caption{WT2}
\label{fig:opt_lw_wt2}
\end{subfigure}

\begin{subfigure}[b]{0.49\textwidth}
\centering
\includegraphics[width=\textwidth]{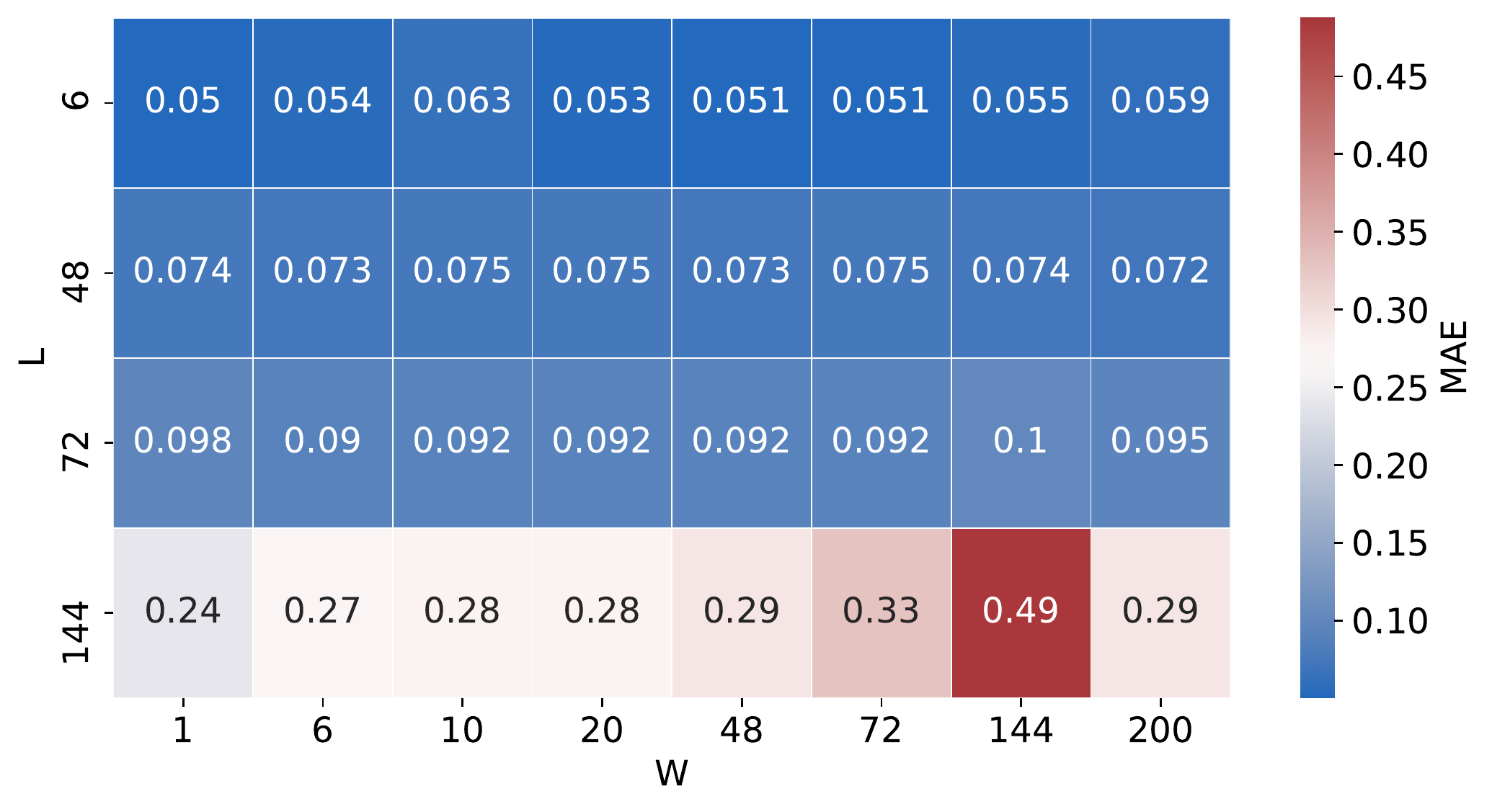}
\caption{WT3}
\label{fig:opt_lw_wt3}
\end{subfigure}
\begin{subfigure}[b]{0.49\textwidth}
\centering
\includegraphics[width=\textwidth]{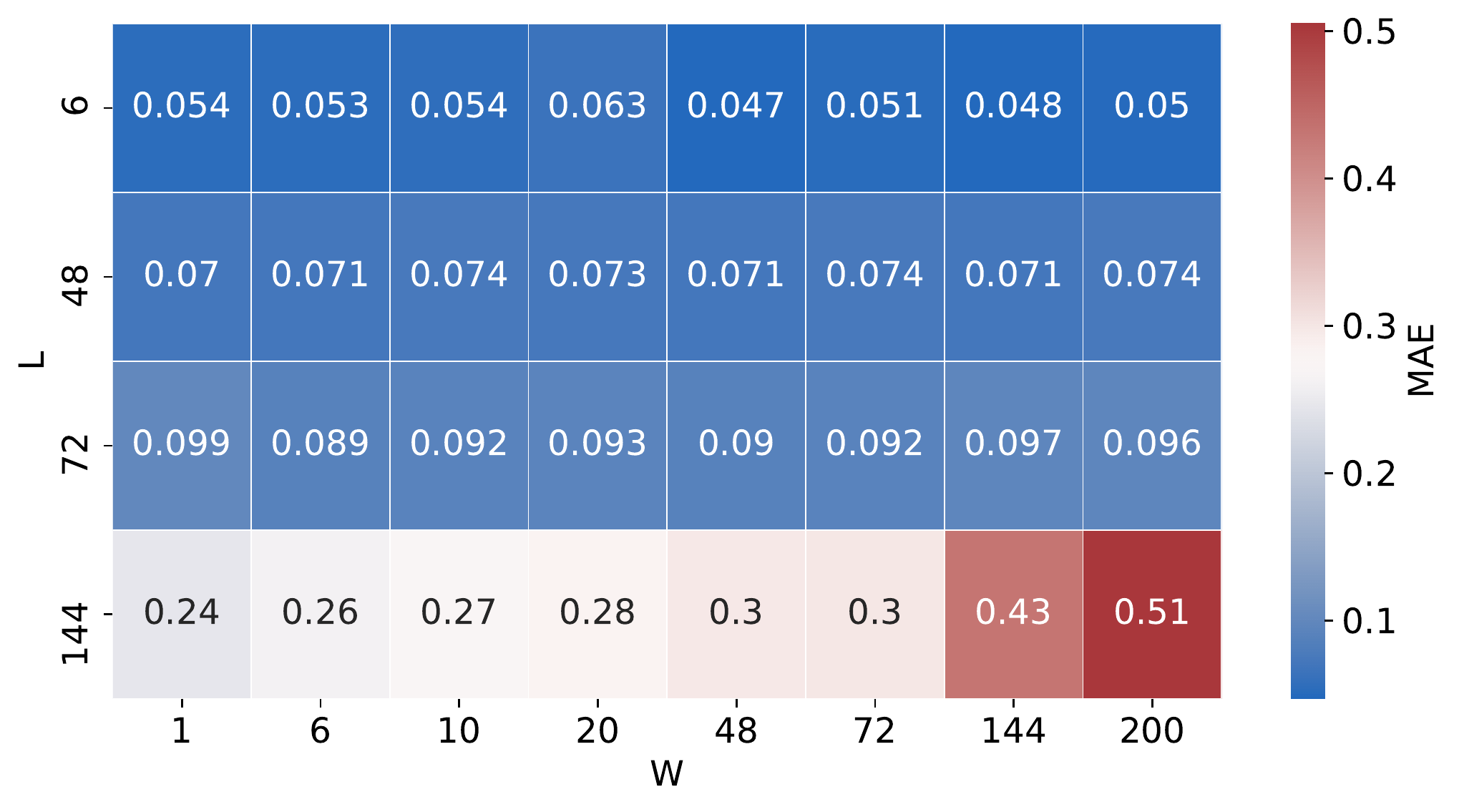}
\caption{WT4}
\label{fig:opt_lw_wt4}
\end{subfigure}
\caption{MAE values obtained by the LSTCN-based model when changing the $w$ and $L$ parameters. As expected, expanding the prediction horizon (that is to say, increasing the number of steps ahead to be predicted) leads to performance degradation of predictions. However, the model does not seem to be especially sensitive to the $w$ parameter, except for larger $L$ values where the error increases as the $w$ gets larger.}
\label{fig:opt_lw}
\end{figure*}

Figure \ref{fig:opt_lw} shows an analysis of the influence of $w$ and $L$ on the model's behavior. The parameters were varied in the discrete set $w=\{1, 6, 10, 20, 48, 72, 144\}$ and $L=\{6, 48, 72, 144\}$, and the MAE computed on the test set was used for comparative purposes. The results did not show a large difference when changing $w$ while keeping $L$ fixed. However, the reduction in model performance was more evident when $L$ increases, which is usual in time series forecasting models.

\begin{figure*}[!h]
\centering

\begin{subfigure}[b]{0.485\textwidth}
\centering
\includegraphics[width=\textwidth]{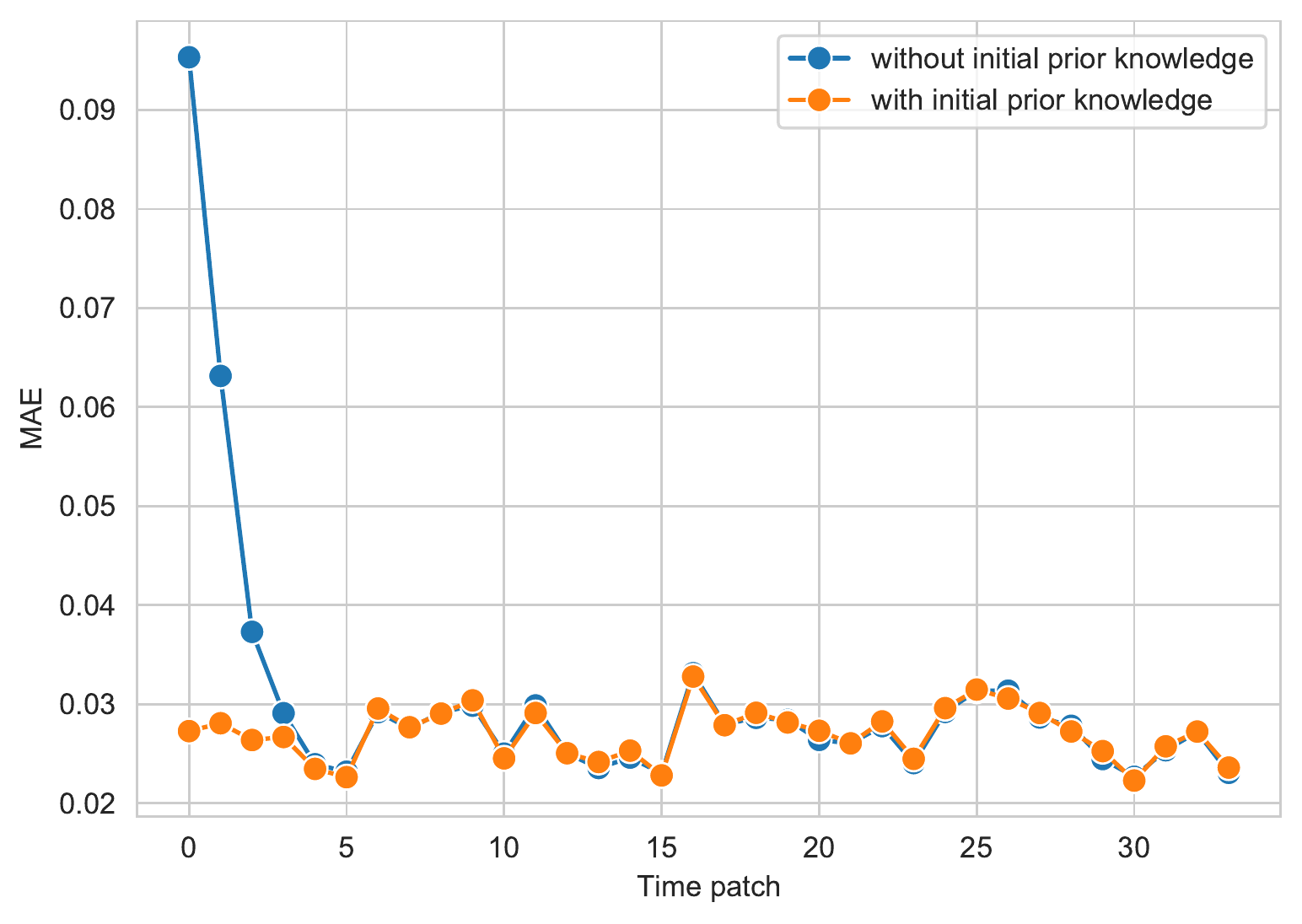}
\caption{WT1}
\label{fig:prior1}
\end{subfigure}
\begin{subfigure}[b]{0.49\textwidth}
\centering
\includegraphics[width=\textwidth]{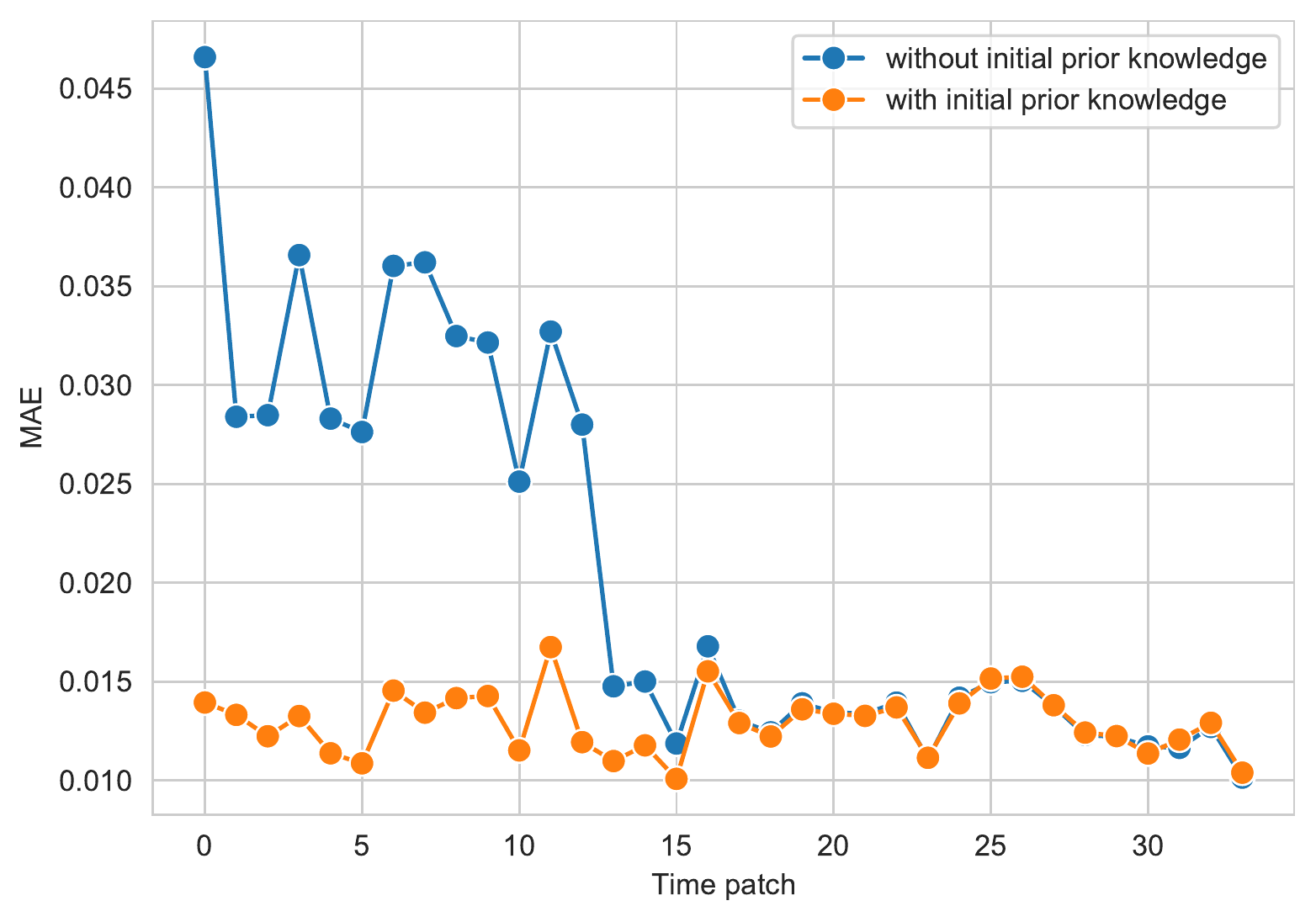}
\caption{WT2}
\label{fig:prior2}
\end{subfigure}

\begin{subfigure}[b]{0.49\textwidth}
\centering
\includegraphics[width=\textwidth]{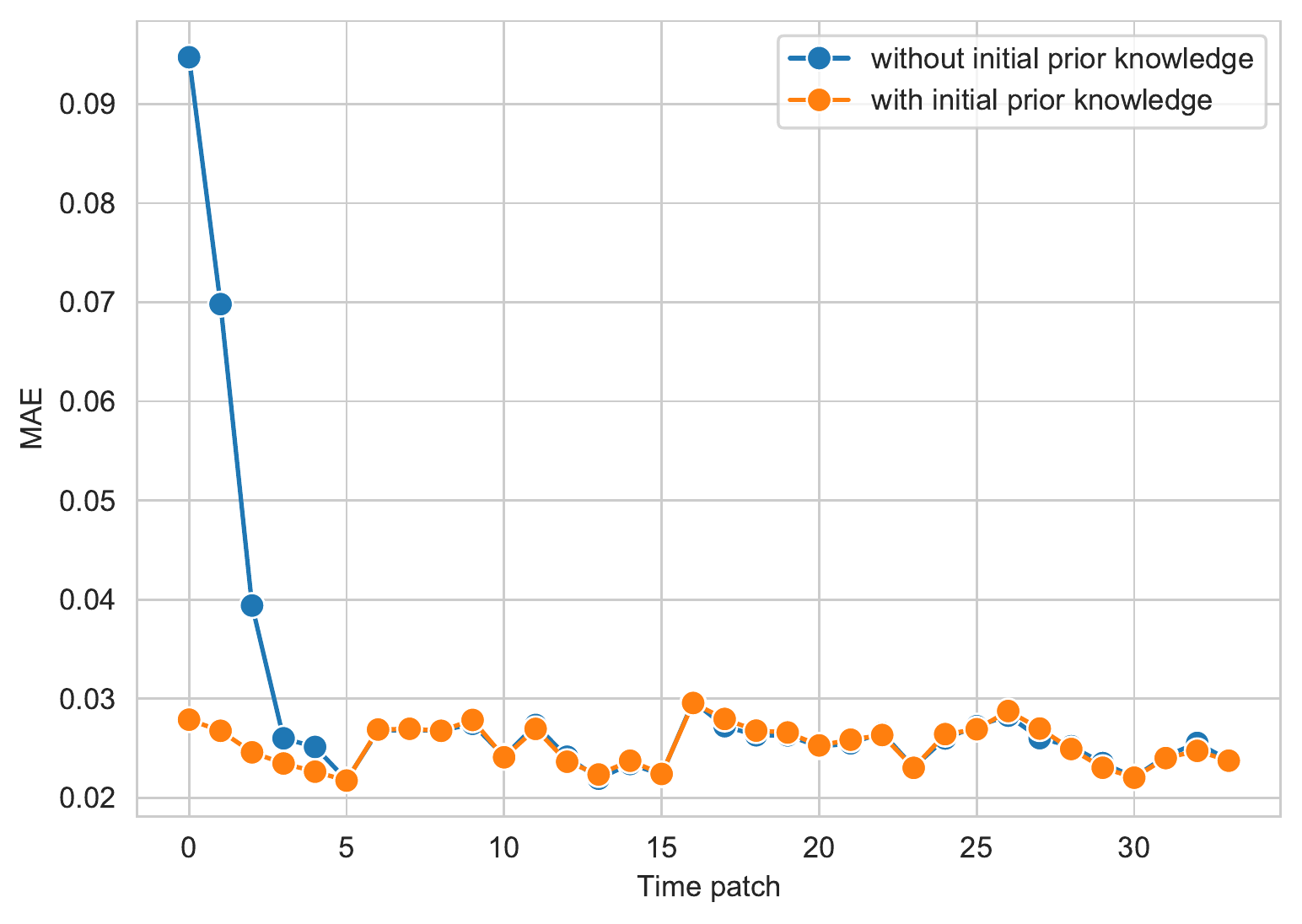}
\caption{WT3}
\label{fig:prior3}
\end{subfigure}
\begin{subfigure}[b]{0.49\textwidth}
\centering
\includegraphics[width=\textwidth]{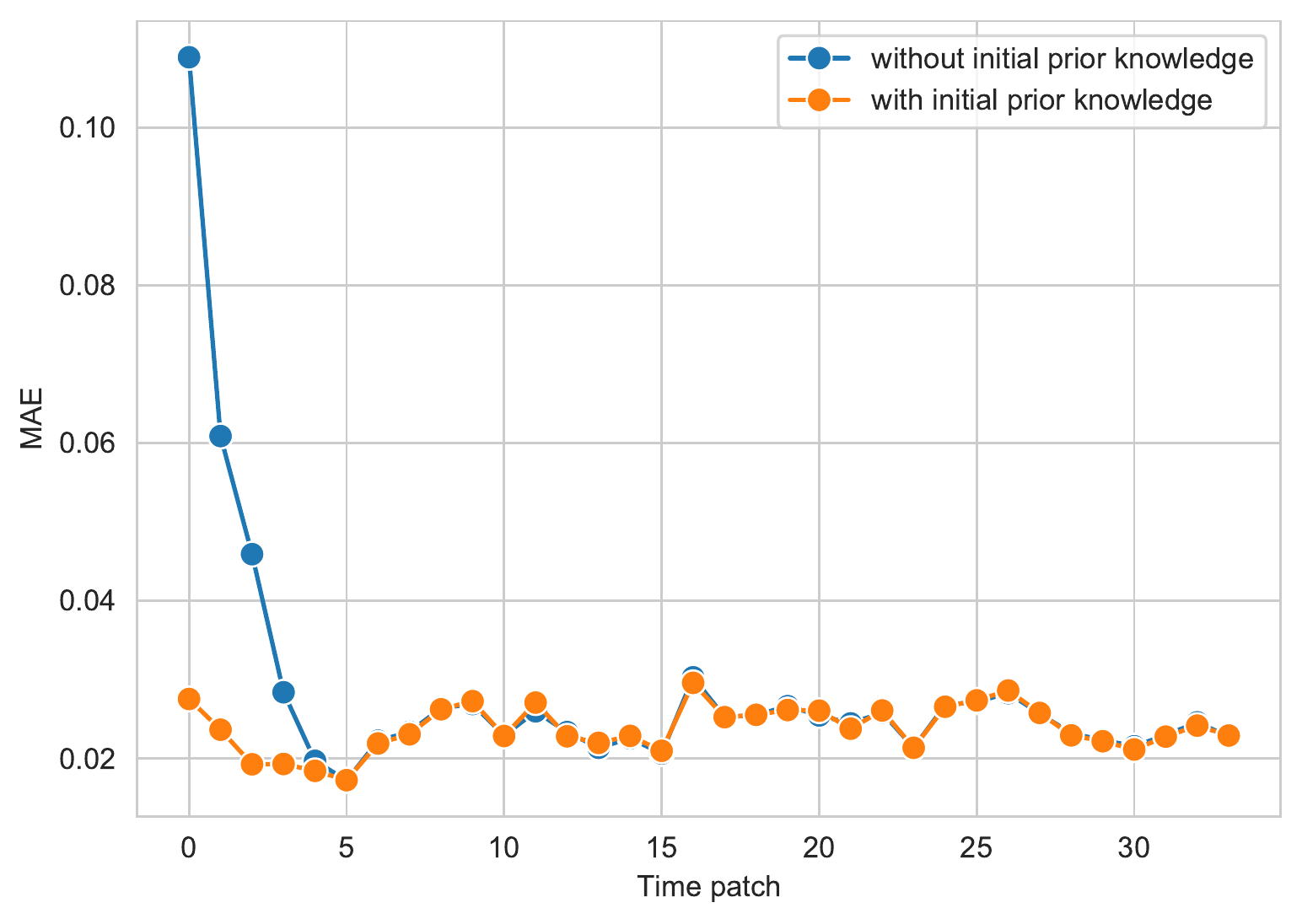}
\caption{WT4}
\label{fig:prior4}
\end{subfigure}
\caption{MAE values obtained by the LSTCN-based model on the four windmill datasets with and without using initial prior knowledge. It can be noticed that the model needs to process more time patches to reduce the error when the model is initialized with a random weight matrix. If this knowledge is not available, the network will still produce good results provided it performs enough iterations.}
\label{fig:MAE_comparison}
\end{figure*}

As mentioned, the knowledge used by the first STCN is extracted from a smoothed representation of the time series data we have. Nevertheless, we can start with a zero-filled matrix if such knowledge is not available. Figure \ref{fig:MAE_comparison} shows the MAE of the predictions in the training set of the four windmills in both settings. Starting from scratch (no knowledge about the data), the LSTCN starts predicting with a large MAE in the first time patch. As new data is received, the network updates its knowledge and reduces the prediction error. In this simulation, we used five time patches such that each STCN block is fitted on the newly received data. The LSTCN model using general knowledge of the time series (assumed as a warm-up) generates small errors from the first time patch.

Tables \ref{tab:mae_results6}, \ref{tab:mae_results48} and \ref{tab:mae_results72} show the results for $L=6, L=48$ and $L=72$, respectively. More explicitly, we report the training and test errors, and the training and test times (in seconds). The LSTCN model obtained the lowest MAE values in all cases (the lowest test error for each windmill is highlighted in boldface). Those results allow us to conclude that our approach is able to produce better forecasting results when compared with well-established recurrent neural networks. It should be noted, however, that such a conclusion is attached to the fact that no hyper-parameter tuning was performed in our simulations.

\begin{table}[h]
\small
\caption{Results for the windmill case study for $L=6$. LSTCN clearly outperforms the other models in both in accuracy and training time.}
\label{tab:mae_results6}
\begin{tabular}{rcrrrr}
\hline
 & \textbf{Model} & \multicolumn{1}{c}{\textbf{\begin{tabular}[c]{@{}c@{}}Training \\ error\end{tabular}}} & \multicolumn{1}{c}{\textbf{\begin{tabular}[c]{@{}c@{}}Test \\ error\end{tabular}}} & \multicolumn{1}{c}{\textbf{\begin{tabular}[c]{@{}c@{}}Training \\ time\end{tabular}}} & \multicolumn{1}{c}{\textbf{\begin{tabular}[c]{@{}c@{}}Test \\ time\end{tabular}}} \\ \hline
\multirow{5}{*}{WT1} & LSTCN & 0.0270 & \textbf{0.0441} & 0.33 & 0.03 \\
 & RNN & 0.1087 & 0.1043 & 17.66 & 0.86 \\
 & LSTM & 0.1036 & 0.0911 & 37.43 & 1.54 \\
 & GRU & 0.1193 & 0.0986 & 45.21 & 1.58 \\
 & HMM & 0.0588 & 0.1219 & 260.27 & 94.88 \\ \hline
\multirow{5}{*}{WT2} & LSTCN & 0.0129 & \textbf{0.0236} & 0.30 & 0.03 \\
 & RNN & 0.1086 & 0.0730 & 20.95 & 1.14 \\
 & LSTM & 0.1035 & 0.0617 & 40.10 & 1.92 \\
 & GRU & 0.1243 & 0.0821 & 37.57 & 1.40 \\
 & HMM & 0.0280 & 0.0942 & 333.61 & 117.57 \\ \hline
\multirow{5}{*}{WT3} & LSTCN & 0.0253 & \textbf{0.0627} & 0.35 & 0.03 \\
 & RNN & 0.1221 & 0.1228 & 17.06 & 0.87 \\
 & LSTM & 0.1107 & 0.1285 & 35.88 & 1.59 \\
 & GRU & 0.1219 & 0.1002 & 37.43 & 1.44 \\
 & HMM & 0.0586 & 0.1878 & 331.04 & 108.29 \\ \hline
\multirow{5}{*}{WT4} & LSTCN & 0.0238 & \textbf{0.0544} & 0.34 & 0.03 \\
 & RNN & 0.1069 & 0.1195 & 17.42 & 1.30 \\
 & LSTM & 0.0979 & 0.0972 & 36.80 & 1.61 \\
 & GRU & 0.1162 & 0.1115 & 35.94 & 1.42 \\
 & HMM & 0.0543 & 0.1665 & 305.35 & 106.25 \\ \hline
\end{tabular}
\end{table}

\begin{table}[h]
\small
\caption{Results for the windmill case study for $L=48$. LSTCN clearly outperforms the other models in both in accuracy and training time.}
\label{tab:mae_results48}
\begin{tabular}{rcrrrr}
\hline
 & \textbf{Model} & \multicolumn{1}{c}{\textbf{\begin{tabular}[c]{@{}c@{}}Training \\ error\end{tabular}}} & \multicolumn{1}{c}{\textbf{\begin{tabular}[c]{@{}c@{}}Test \\ error\end{tabular}}} & \multicolumn{1}{c}{\textbf{\begin{tabular}[c]{@{}c@{}}Training \\ time\end{tabular}}} & \multicolumn{1}{c}{\textbf{\begin{tabular}[c]{@{}c@{}}Test \\ time\end{tabular}}} \\ \hline
\multirow{5}{*}{WT1} & LSTCN & 0.0383 & \textbf{0.0751} & 0.79 & 0.04 \\
 & RNN & 0.5963 & 0.6489 & 76.38 & 3.35 \\
 & LSTM & 0.1248 & 0.1048 & 869.13 & 20.26 \\
 & GRU & 0.1641 & 0.1086 & 762.64 & 13.42 \\
 & HMM & 0.0794 & 0.1378 & 1585.80 & 445.12 \\ \hline
\multirow{5}{*}{WT2} & LSTCN & 0.0190 & \textbf{0.0407} & 0.77 & 0.04 \\
 & RNN & 0.5340 & 0.4560 & 134.81 & 3.80 \\
 & LSTM & 0.1424 & 0.0777 & 906.91 & 20.49 \\
 & GRU & 0.2317 & 0.0821 & 625.43 & 11.74 \\
 & HMM & 0.0388 & 0.0821 & 1586.91 & 441.19 \\ \hline
\multirow{5}{*}{WT3} & LSTCN & 0.0371 & \textbf{0.0750} & 0.87 & 0.04 \\
 & RNN & 0.4991 & 0.5569 & 100.70 & 3.46 \\
 & LSTM & 0.1332 & 0.1168 & 915.26 & 20.15 \\
 & GRU & 0.1887 & 0.1100 & 645.27 & 11.18 \\
 & HMM & 0.0791 & 0.1582 & 1462.68 & 413.93 \\ \hline
\multirow{5}{*}{WT4} & LSTCN & 0.0351 & \textbf{0.0744} & 1.08 & 0.05 \\
 & RNN & 0.3447 & 0.2579 & 169.40 & 3.43 \\
 & LSTM & 0.1254 & 0.1060 & 888.81 & 20.02 \\
 & GRU & 0.1634 & 0.1107 & 662.83 & 11.45 \\
 & HMM & 0.0744 & 0.1327 & 1791.60 & 513.98 \\ \hline
\end{tabular}
\end{table}

\begin{table}[!h]
\small
\caption{Results for the windmill case study for $L=72$. LSTCN clearly outperforms the other models in both in accuracy and training time.}
\label{tab:mae_results72}
\begin{tabular}{rcrrrr}
\hline
 & \textbf{Model} & \multicolumn{1}{c}{\textbf{\begin{tabular}[c]{@{}c@{}}Training \\ error\end{tabular}}} & \multicolumn{1}{c}{\textbf{\begin{tabular}[c]{@{}c@{}}Test \\ error\end{tabular}}} & \multicolumn{1}{c}{\textbf{\begin{tabular}[c]{@{}c@{}}Training \\ time\end{tabular}}} & \multicolumn{1}{c}{\textbf{\begin{tabular}[c]{@{}c@{}}Test \\ time\end{tabular}}} \\ \hline
\multirow{5}{*}{WT1} & LSTCN & 0.0370 & \textbf{0.0916} & 0.77 & 0.05 \\
 & RNN & 0.5187 & 0.5350 & 107.74 & 5.29 \\
 & LSTM & 0.1376 & 0.1237 & 1516.66 & 39.05 \\
 & GRU & 0.1891 & 0.1250 & 1852.98 & 29.47 \\
 & HMM & 0.0832 & 0.1434 & 2070.78 & 594.75 \\ \hline
\multirow{5}{*}{WT2} & LSTCN & 0.0184 & \textbf{0.0479} & 1.02 & 0.05 \\
 & RNN & 0.6490 & 0.6317 & 149.05 & 6.79 \\
 & LSTM & 0.1320 & 0.1021 & 1483.30 & 38.94 \\
 & GRU & 0.2304 & 0.1641 & 1981.13 & 31.73 \\
 & HMM & 0.0409 & 0.0817 & 2243.35 & 651.34 \\ \hline
\multirow{5}{*}{WT3} & LSTCN & 0.0363 & \textbf{0.0925} & 0.82 & 0.04 \\
 & RNN & 0.5524 & 0.6116 & 127.70 & 6.70 \\
 & LSTM & 0.1441 & 0.1360 & 1471.54 & 36.88 \\
 & GRU & 0.1846 & 0.1468 & 1820.24 & 30.58 \\
 & HMM & 0.0827 & 0.1590 & 2096.07 & 601.62 \\ \hline
\multirow{5}{*}{WT4} & LSTCN & 0.0355 & \textbf{0.0916} & 1.10 & 0.05 \\
 & RNN & 0.5413 & 0.5911 & 115.70 & 5.55 \\
 & LSTM & 0.1358 & 0.1202 & 1453.89 & 38.38 \\
 & GRU & 0.1813 & 0.1178 & 1737.80 & 33.77 \\
 & HMM & 0.0784 & 0.1330 & 2270.64 & 634.80 \\ \hline
\end{tabular}
\end{table}


Another clear advantage of LSTCN over these state-of-the-art algorithms is the reduced training and test times. Re-training the model quickly when a new piece of data arrives while retaining the knowledge we have learned so far is a key challenge in online learning settings. Recurrent neural models such as RNN, LSTM, GRU use a backpropagation-based learning algorithm to compute the weights regulating the network behavior. The algorithm needs to iterate multiple times over the data with limited vectorization possibilities.

Overall, there is a trade-off between accuracy and training time when it comes to the batch size. The smaller the batch size in the backpropagation learning, the more accurate the predictions are expected to be. However, smaller batch sizes make the training process slower. Another issue with gradient-based optimization methods is that they usually operate in a stochastic fashion, thus making them quite sensitive to the initial conditions. Notice that HMM also requires several iterations to build the probability transition matrix.

\begin{figure}[!htbp]
\centering
\begin{subfigure}[b]{0.48\textwidth}
\includegraphics[width=\textwidth]{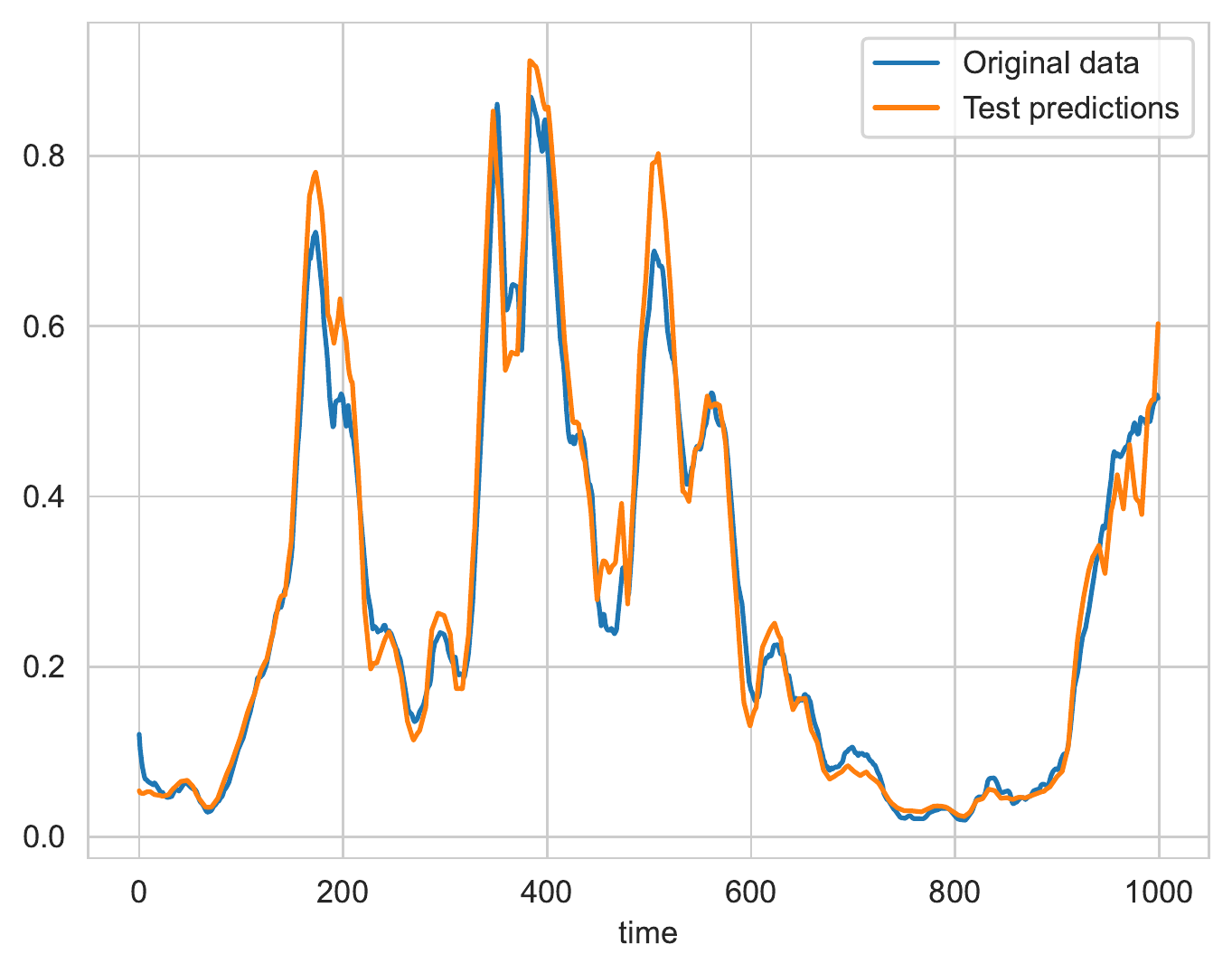}
\caption{$L=6$}
\label{fig:WT1L6}
\end{subfigure}
\hfill
\begin{subfigure}[b]{0.48\textwidth}
\includegraphics[width=\textwidth]{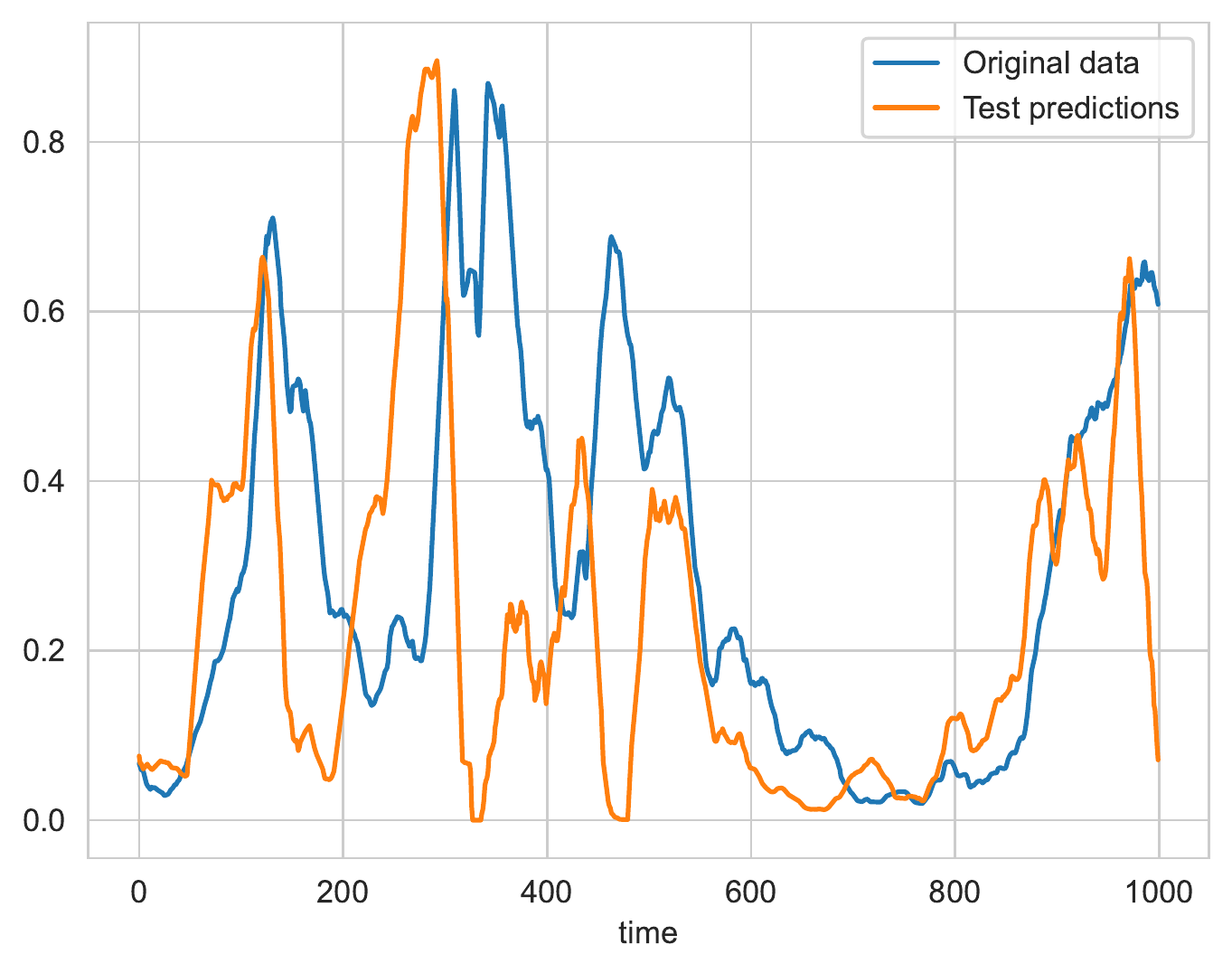}
\caption{$L=48$}
\label{fig:WT1L48}
\end{subfigure}

\begin{subfigure}[b]{0.48\textwidth}
\centering
\includegraphics[width=\textwidth]{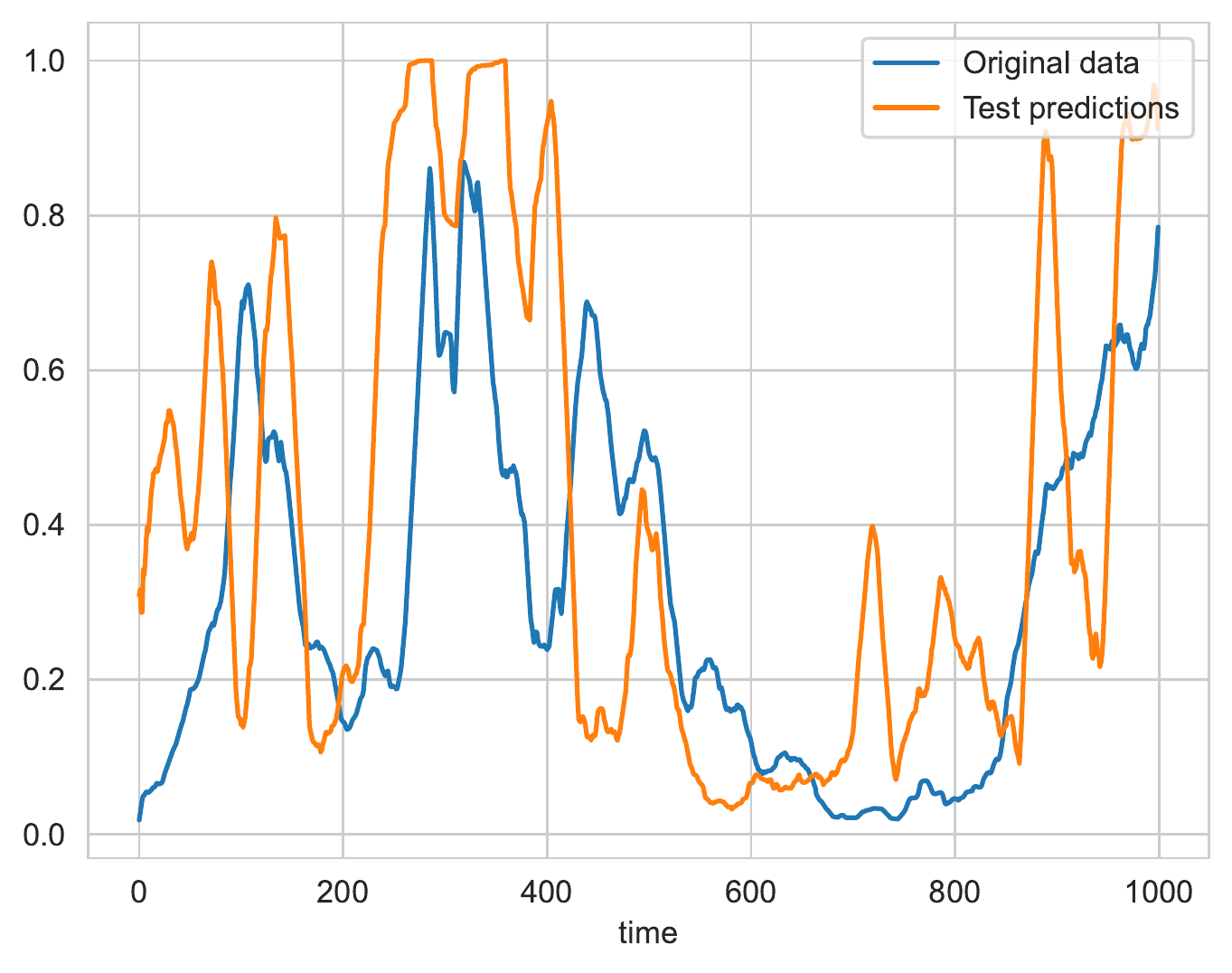}
\caption{$L=72$}
\label{fig:WT1L72}
\end{subfigure}

\caption{Moving average power predictions ($w=24$) for the first windmill with (a) $L=6$, (b) $L=48$ and (c) $L=72$.}
\label{fig:WT1}
\end{figure}

\begin{figure}[!htbp]
\centering
\begin{subfigure}[b]{0.48\textwidth}
\includegraphics[width=\textwidth]{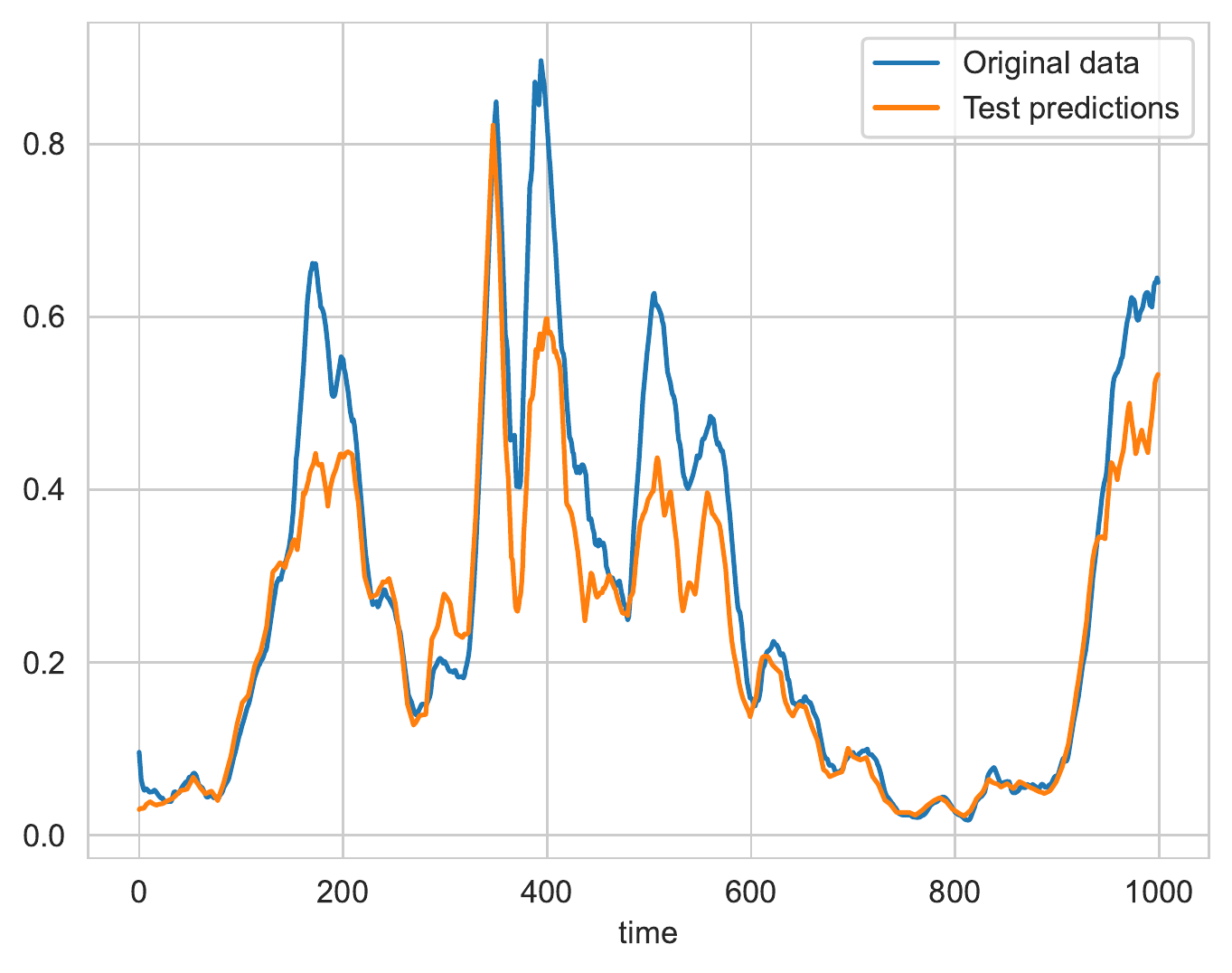}
\caption{$L=6$}
\label{fig:WT2L6}
\end{subfigure}
\hfill
\begin{subfigure}[b]{0.48\textwidth}
\includegraphics[width=\textwidth]{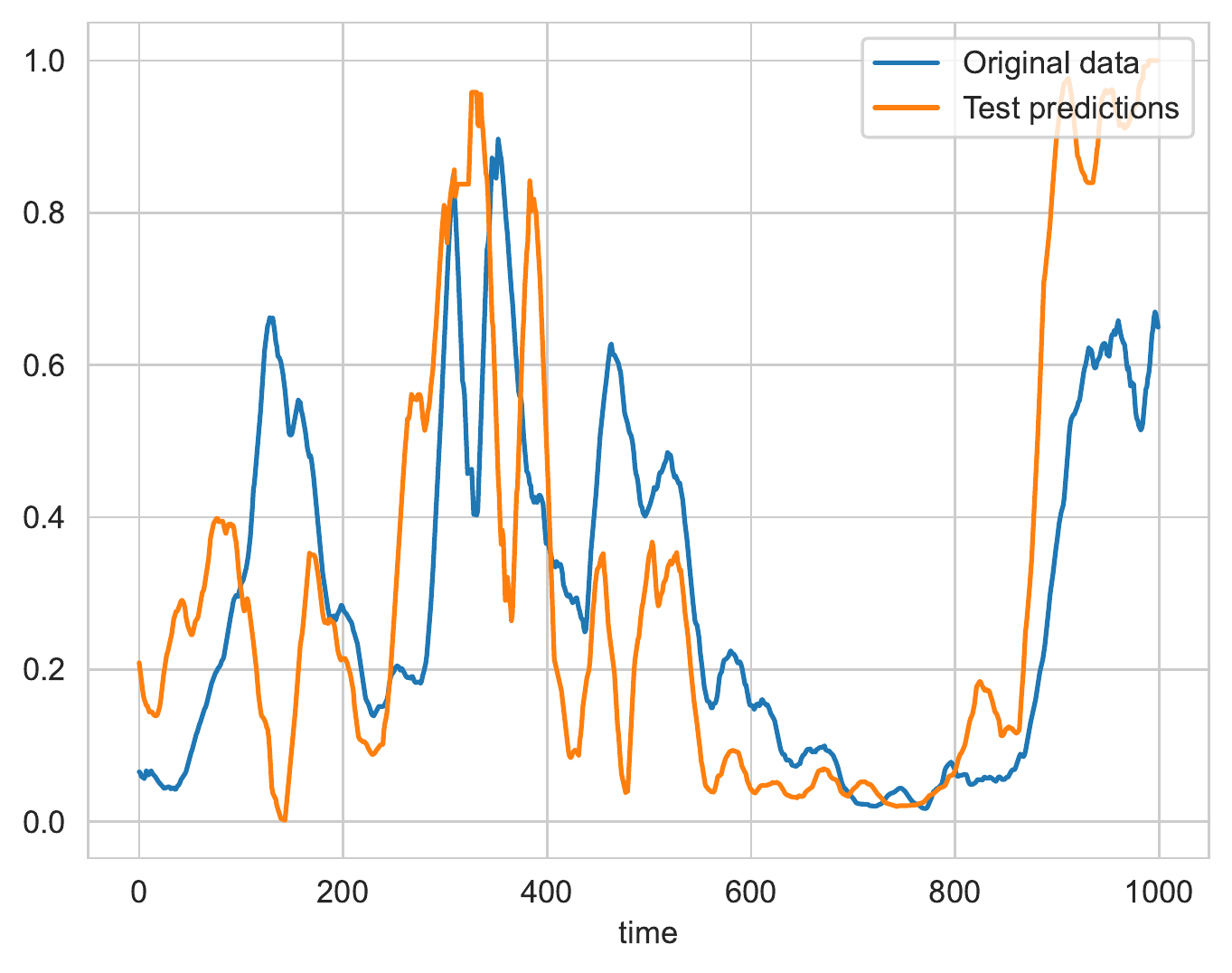}
\caption{$L=48$}
\label{fig:WT2L48}
\end{subfigure}

\begin{subfigure}[b]{0.48\textwidth}
\centering
\includegraphics[width=\textwidth]{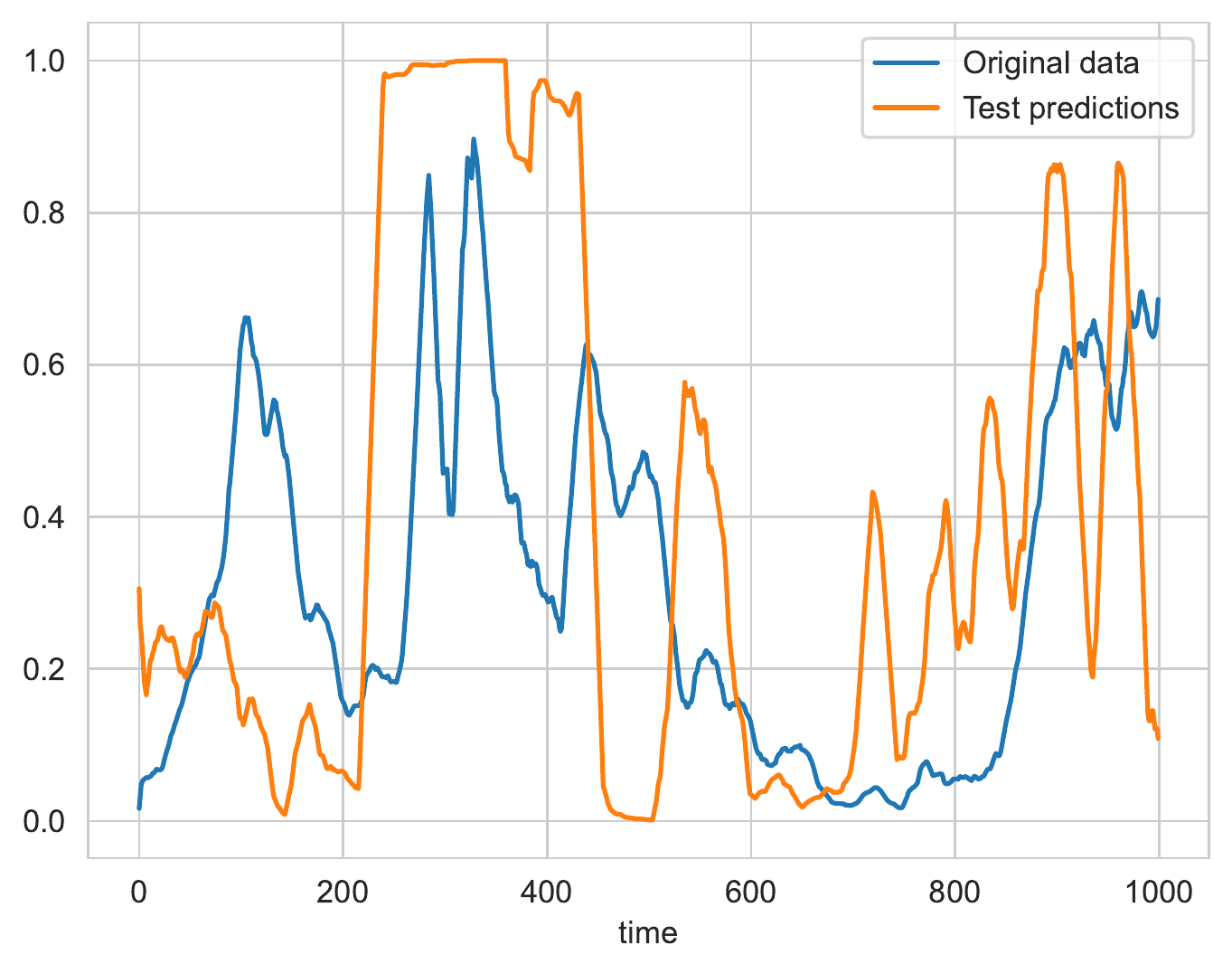}
\caption{$L=72$}
\label{fig:WT2L72}
\end{subfigure}

\caption{Moving average power predictions ($w=24$) for the second windmill with (a) $L=6$, (b) $L=48$ and (c) $L=72$.}
\label{fig:WT2}
\end{figure}


\begin{figure}[!htbp]
\centering
\begin{subfigure}[b]{0.48\textwidth}
\includegraphics[width=\textwidth]{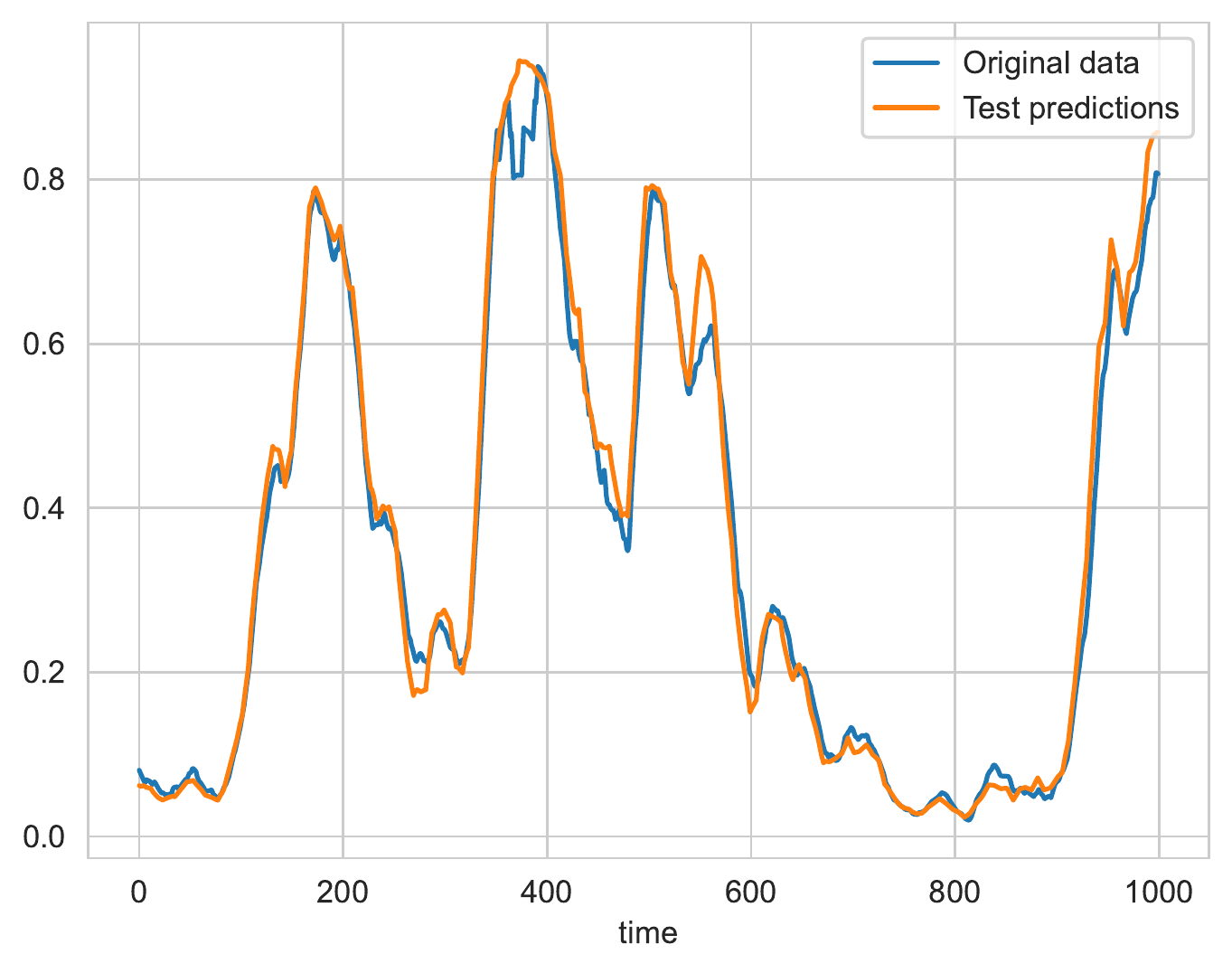}
\caption{$L=6$}
\label{fig:WT3L6}
\end{subfigure}
\hfill
\begin{subfigure}[b]{0.48\textwidth}
\includegraphics[width=\textwidth]{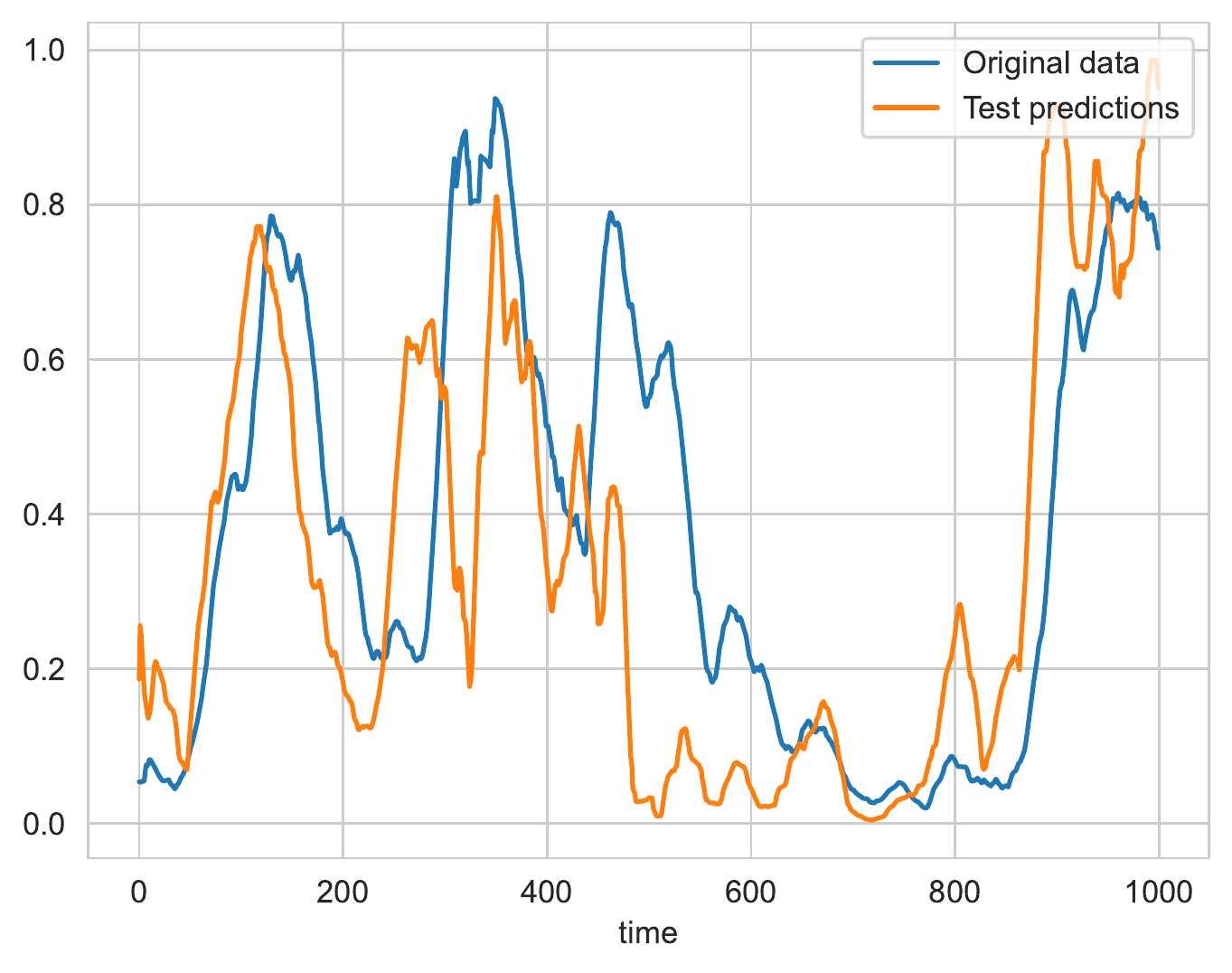}
\caption{$L=48$}
\label{fig:WT3L48}
\end{subfigure}

\begin{subfigure}[b]{0.48\textwidth}
\centering
\includegraphics[width=\textwidth]{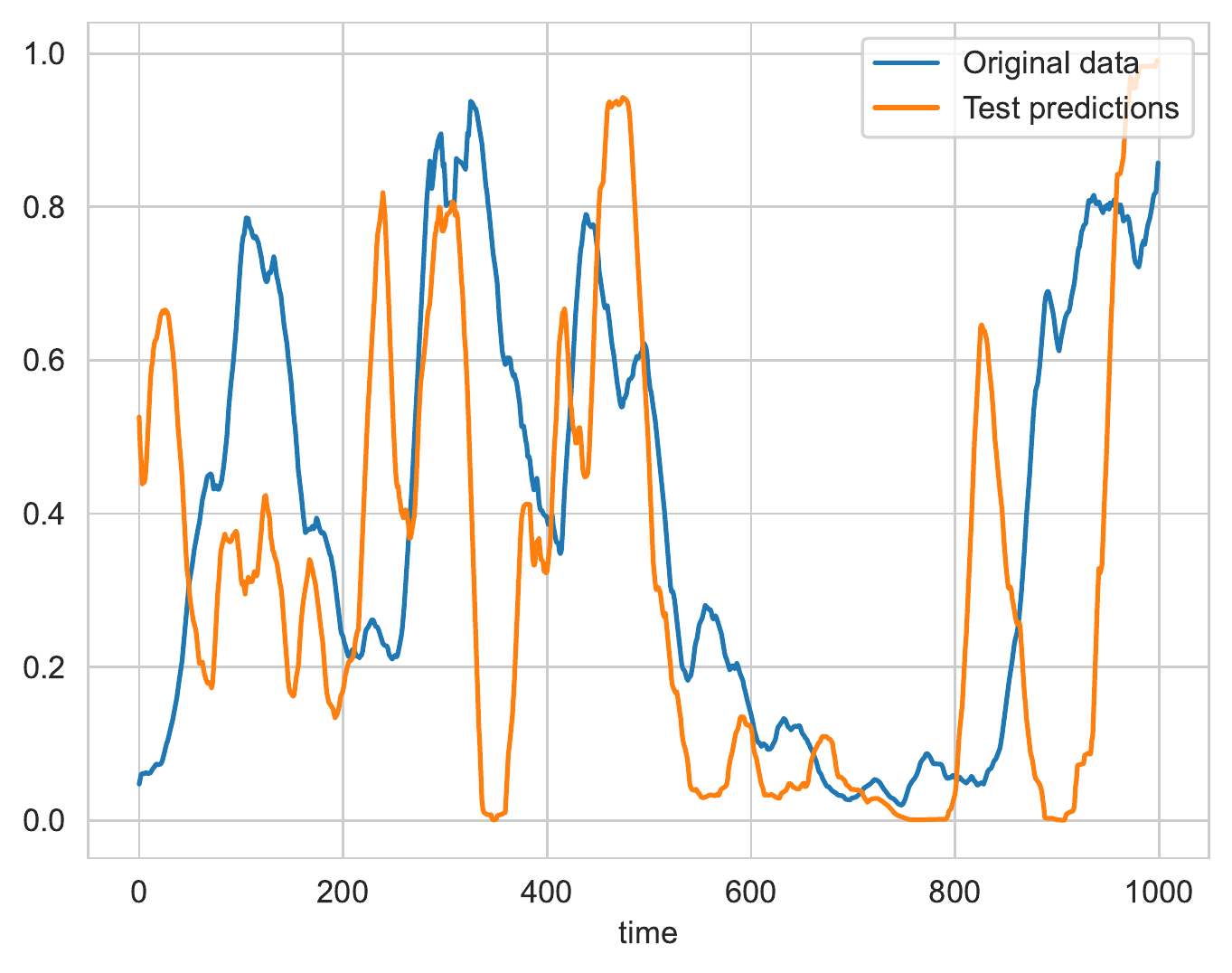}
\caption{$L=72$}
\label{fig:WT3L72}
\end{subfigure}

\caption{Moving average power predictions ($w=24$) for the third windmill with (a) $L=6$, (b) $L=48$ and (c) $L=72$.}
\label{fig:WT3}
\end{figure}


\begin{figure}[!htbp]
\centering
\begin{subfigure}[b]{0.48\textwidth}
\includegraphics[width=\textwidth]{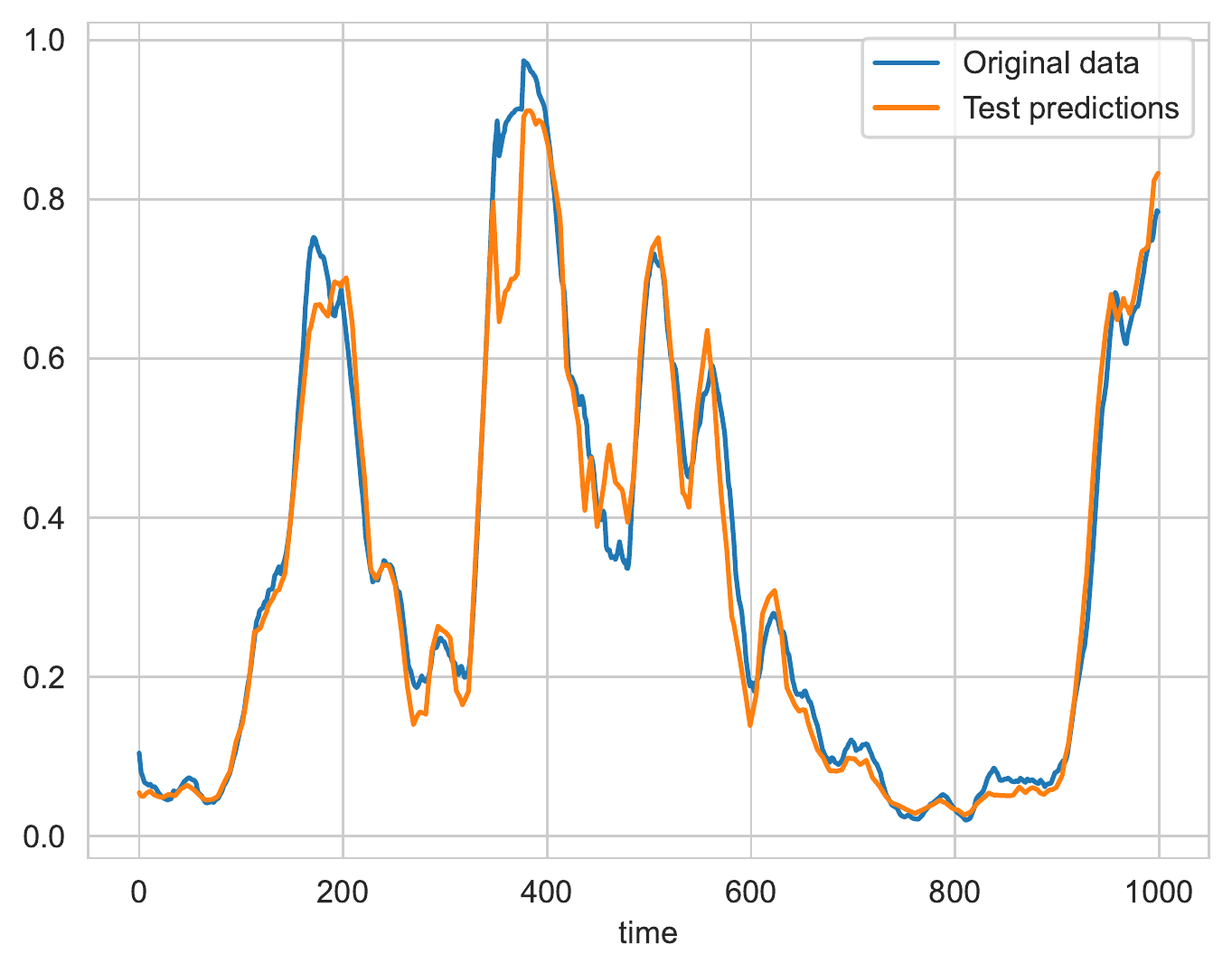}
\caption{$L=6$}
\label{fig:WT4L6}
\end{subfigure}
\hfill
\begin{subfigure}[b]{0.48\textwidth}
\includegraphics[width=\textwidth]{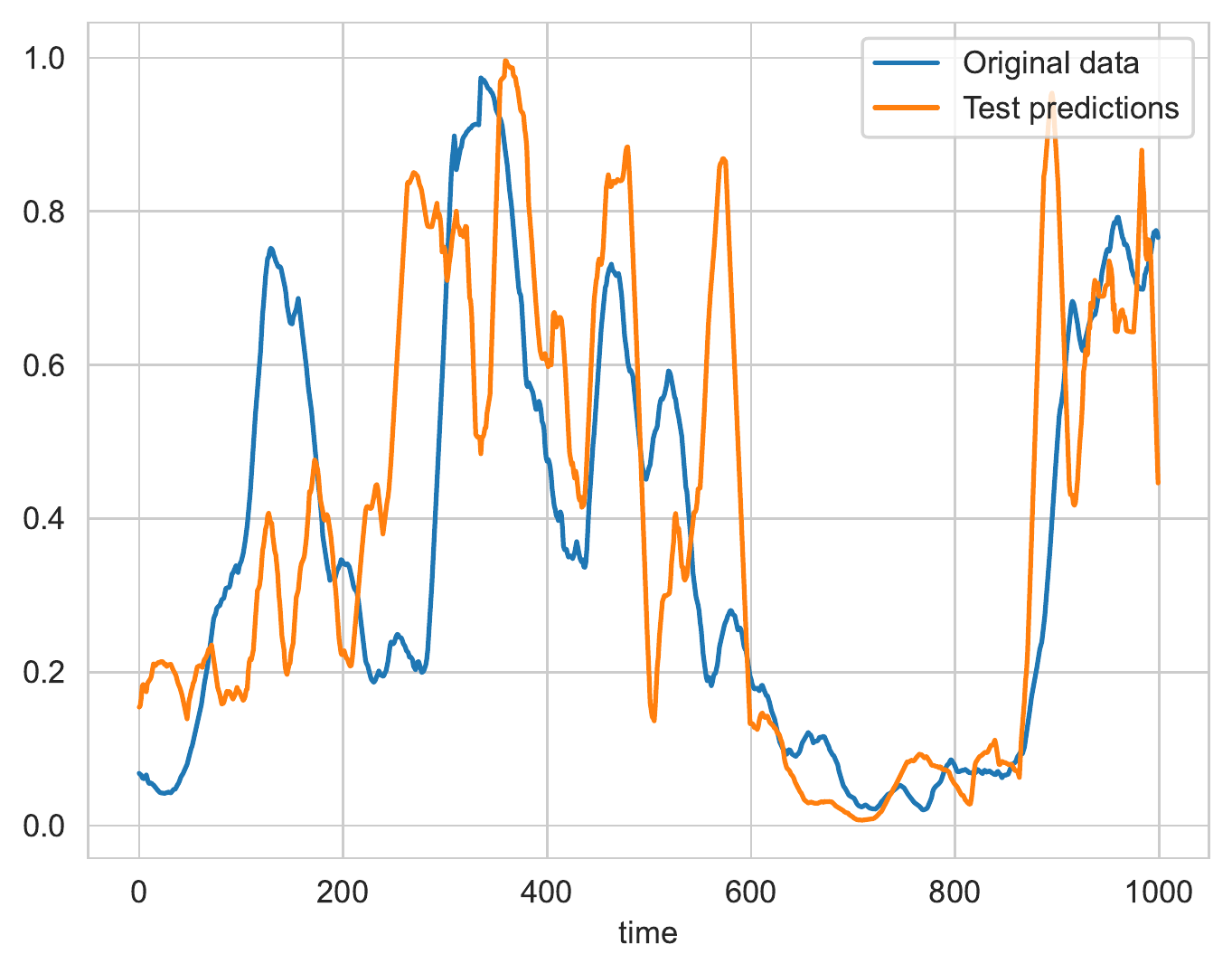}
\caption{$L=48$}
\label{fig:WT4L48}
\end{subfigure}

\begin{subfigure}[b]{0.48\textwidth}
\centering
\includegraphics[width=\textwidth]{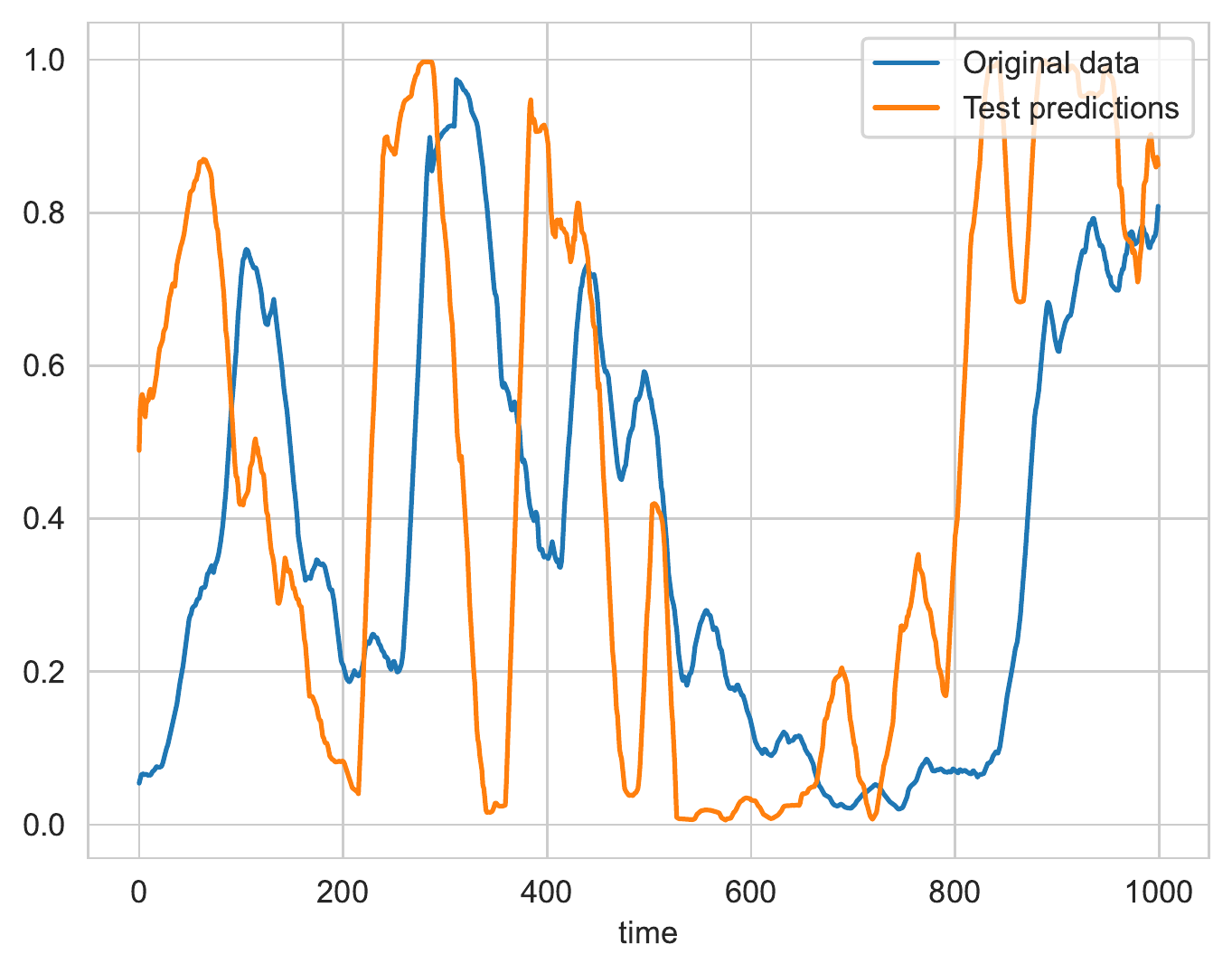}
\caption{$L=72$}
\label{fig:WT4L72}
\end{subfigure}

\caption{Moving average power predictions ($w=24$) for the fourth windmill with (a) $L=6$, (b) $L=48$ and (c) $L=72$.}
\label{fig:WT4}
\end{figure}

To alleviate the problem caused by larger prediction horizons, we could increase the batch size in our model (or decrease the batch size of other recurrent models used for comparison purposes). In that way, we will have more data in each training process, which will likely lead to models with improved predictive power. Alternatively, we could adopt an incremental learning approach to reuse data concerning previous time patches as defined by a given window parameter. However, we should be aware that many online learning problems operate on volatile data that is just available for a short period.

\section{Concluding remarks}
\label{sec:remarks}

In this paper, we investigated the performance of Long Short-term Cognitive Networks to forecast windmill time series in online setting scenarios. This brand-new recurrent model system consists of a sequence of Short-term Cognitive Network blocks. Each of these blocks is trained with the available data at that moment in time such that the learned knowledge is propagated to the next blocks. Therefore, the network is able to adjust its knowledge to new information, which makes this model suitable for online settings since we retain the knowledge learned from previous learning processes.

The experiments conducted using four windmill datasets reported that our approach outperforms other state-of-the-art recurrent neural networks in terms of MAE. In addition, the proposed LSTCN-based model is significantly faster than these recurrent models when it comes to both training and test times. Such a feature is of paramount relevance when designing forecasting models operating in online learning modes. Regrettably, the overall performance of all forecasting models deteriorated when increasing the number of steps ahead to be predicted. While this result is not surprising, further efforts are needed to build forecasting models with better scalability properties as defined by the prediction horizon.

Before concluding our paper, it is worth mentioning that the proposed architecture for online time series forecasting is not restricted to windmill data. Instead, the architecture can be applied to any univariate or multivariate time series provided that the proper pre-processing steps are conducted.

\section*{Acknowledgement}
Alejandro Morales and Koen Vanhoof from Hasselt University would like to thank the support received by the Flanders AI Research Program, as well as other partners involved in this project. Agnieszka Jastrzebska’s contribution was founded by the POB Research Center for Artificial Intelligence and Robotics of Warsaw University of Technology within the Excellence Initiative Program - Research University (ID-UB). The authors would like to thank Isel Grau from the Eindhoven University of Technology for revising the paper.



\end{document}